\documentclass[10pt,twocolumn,letterpaper]{article}

\usepackage[pagenumbers]{style}

\usepackage{graphicx}
\usepackage{amsmath}
\usepackage{amssymb}
\usepackage{booktabs}
\usepackage{algorithm}
\usepackage{listings}
\usepackage[accsupp]{axessibility}  

\usepackage[pagebackref,breaklinks,colorlinks]{hyperref}
\usepackage{bbm}
\usepackage{cancel}
\usepackage{multirow}
\usepackage{enumitem}
\usepackage{makecell}
\usepackage{subcaption}
\usepackage{marvosym}

\usepackage[square,sort,comma,numbers]{natbib}
\usepackage{multibib}
\newcites{appendix}{Appendix References}

\usepackage[capitalize]{cleveref}
\crefname{section}{Sec.}{Secs.}
\Crefname{section}{Section}{Sections}
\Crefname{table}{Table}{Tables}
\crefname{table}{Tab.}{Tabs.}

\newcommand{\name}{UniGrad}
\allowdisplaybreaks

\makeatletter
\def\blfootnote{\xdef\@thefnmark{}\@footnotetext}
\makeatother

\begin{document}

\title{Exploring the Equivalence of Siamese Self-Supervised Learning via\\ A Unified Gradient Framework}

\vspace{-0.5em}
\author{
Chenxin Tao$^{1*\dag}$,
Honghui Wang$^{1*\dag}$,
Xizhou Zhu$^{2*}$,
Jiahua Dong$^{3*\dag}$,\\
Shiji Song$^1$,
Gao Huang$^{1,4}$,
Jifeng Dai$^{2,4}$\textsuperscript{\Letter}\\
$^1$Tsinghua University, $^2$SenseTime Research, $^3$Zhejiang University\\
$^4$Beijing Academy of Artificial Intelligence, Beijing, China\\
{\tt\small \{tcx20, wanghh20\}@mails.tsinghua.edu.cn, \{zhuwalter, daijifeng\}@sensetime.com}\\
{\tt\small cnjiahuadong@gmail.com, shijis@mail.tsinghua.edu.cn, gaohuang@tsinghua.edu.cn}
}
\maketitle

\begin{abstract}
Self-supervised learning has shown its great potential to extract powerful visual representations without human annotations. Various works are proposed to deal with self-supervised learning from different perspectives: (1) contrastive learning methods (e.g., MoCo, SimCLR) utilize both positive and negative samples to guide the training direction; (2) asymmetric network methods (e.g., BYOL, SimSiam) get rid of negative samples via the introduction of a predictor network and the stop-gradient operation; (3) feature decorrelation methods (e.g., Barlow Twins, VICReg) instead aim to reduce the redundancy between feature dimensions. These methods appear to be quite different in the designed loss functions from various motivations. The final accuracy numbers also vary, where different networks and tricks are utilized in different works. In this work, we demonstrate that these methods can be unified into the same form. Instead of comparing their loss functions, we derive a unified formula through gradient analysis. Furthermore, we conduct fair and detailed experiments to compare their performances. It turns out that there is little gap between these methods, and the use of momentum encoder is the key factor to boost performance.

From this unified framework, we propose \name, a simple but effective gradient form for self-supervised learning. It does not require a memory bank or a predictor network, but can still achieve state-of-the-art performance and easily adopt other training strategies. Extensive experiments on linear evaluation and many downstream tasks also show its effectiveness. Code is released at \url{https://github.com/fundamentalvision/UniGrad}.
\end{abstract}

\blfootnote{\noindent $^{*}$Equal contribution. $^{\dag}$This work is done when Chenxin Tao, Honghui Wang, and Jiahua Dong are interns at SenseTime Research. \textsuperscript{\Letter}Corresponding author.}

\begin{figure}[t]
    \centering
    \vspace{-1.0em}
    \includegraphics[width=0.4\textwidth]{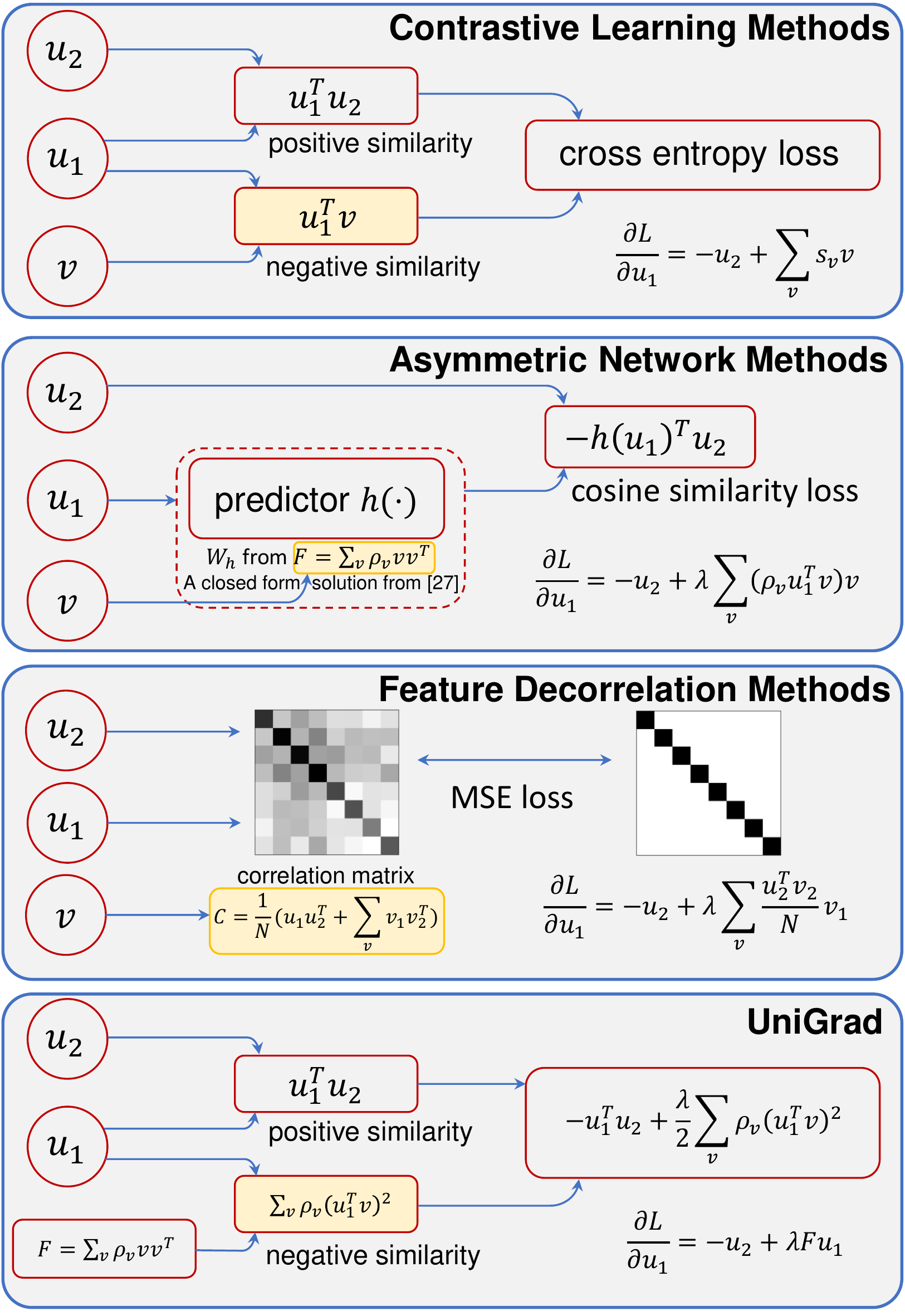}
    \vspace{-0.5em}
    \caption{Overview of three typical types of self-supervised learning methods and our proposed \name. $u_1$ and $u_2$ are two augmented views of the same image. $v$ denote views of other images. We find these methods have a similar gradient structure composed of the positive and negative gradients, which can be analogous to positive and negative samples in contrastive learning. Because some methods do not explicitly utilize negative samples, we highlight the source of negative gradient in each method.}
    \label{fig:overview}
    \vspace{-1.5em}
\end{figure}

\section{Introduction}
\label{sec:intro}
Self-supervised learning (SSL) has recently attracted much research interest~\cite{he2020momentum, chen2020simple, jean-bastien2020bootstrap, chen2021exploring, zbontar2021barlow, bardes2021vicreg}. It has shown the potential to extract powerful visual representations that are competitive with supervised learning, and delivered superior performance on multiple visual tasks.

Recent works deal with SSL from different points of view, leading to three typical types of methods (see Figure~\ref{fig:overview}), while siamese networks are always employed.
\textit{Contrastive Learning} methods~\cite{he2020momentum, chen2020improved, chen2021empirical, chen2020simple} aim to reduce the distance between two augmented views from the same image (positive samples), and push apart views from different images (negative samples). Negative samples play an important role in these methods to avoid representational collapse.
\textit{Asymmetric Network} methods~\cite{jean-bastien2020bootstrap, chen2021exploring} claim that only adopting positive samples is sufficient. The key is the introduction of asymmetric network architecture. In these methods, a predictor network is only appended after one branch of the siamese network, and the other branch is detached from the gradient back-propagation. Although these methods have achieved impressive performance, they are still poorly understood. Recent work~\cite{tian2021understanding} has tried to analyze their training dynamics, but still lacks a straight-forward explanation. 
\textit{Feature Decorrelation} methods~\cite{ermolov2021whitening, zbontar2021barlow, bardes2021vicreg, hua2021feature} have recently been proposed as a new solution for SSL. They instead focus on reducing the redundancy between different feature dimensions.
These methods seem to be highly different in how to learn representations, and it's also hard to compare their performances because different networks and tricks are utilized in different works. With so many different methods, it is natural to ask: What is the relationship among them? Are there any connections among the working mechanisms behind them? What factors actually cause the performance difference?

In this work, we unify the aforementioned three typical types of SSL methods in a unified framework. Instead of comparing their loss functions, the unified formula is derived through gradient analysis. We find that all these methods have similar gradient formulas. They consist of three components: the positive gradient, the negative gradient and a scalar that balances these two terms. 
The positive gradient is the representation of another augmented view from the same image, while the negative gradient is a weighted combination of the representations from different images. The effects of these two terms are similar to that of positive and negative samples in contrastive learning methods. This suggests that these methods share a similar working mechanism, but organize the loss functions in different manners.
Moreover, since these methods are different in the specific formula of the gradient, we conduct fair and detailed experiments for comparison. It turns out that different gradient formulas result in close performance, and what really matters is the use of momentum encoder.

From this unified framework, we propose a concise but effective gradient formula named \name, which explicitly maximizes the similarity between positive samples and expects the similarity between negative samples to be zero.
This formula does not require memory bank or asymmetric network, and can easily adopt prevalent augmentation strategies (\eg, CutMix~\cite{yun2019cutmix} and multi-crop~\cite{caron2020unsupervised,caron2021emerging}) to further improve the performance. Extensive experiments show that our method is competitive on various tasks, including the standard linear evaluation protocol, the semi-supervised learning task and various downstream vision tasks. Our contribution can be summarized as:
\begin{itemize}[leftmargin=2em]
    \vspace{-0.5em}
    \item A unified framework is proposed for different self-supervised learning methods through the perspective of gradient analysis. This shows that although previous works seem to be distinct in loss functions, they actually work in a similar mechanism;
    \vspace{-0.5em}
    \item Different self-supervised learning methods are compared under a fair and controlled experiment setting. The results show that they can achieve similar performance, while the momentum encoder is actually the key factor affecting the final performance;
    \vspace{-0.5em}
    \item \name\ is proposed as a concise but effective gradient formula for self-supervised learning. Extensive experiments demonstrate its competitive performance.
\end{itemize}

\section{Related Work}
\label{sec:related}

\noindent\textbf{Contrastive Learning Methods} have long been studied in the area of self-supervised learning~\cite{bromley1993signature, chopra2005learning, weinberger2009distance}. The main idea is to discriminate positive and negative samples.
Since the propose of InfoNCE~\cite{oord2018representation}, many recent works~\cite{he2020momentum,chen2020simple,chen2020improved, cao2020parametric,kalantidis2020hard,chen2021empirical,zhu2021improving,li2020prototypical,caron2020unsupervised,hu2021adco} have pushed the performance to a new height. In these methods, negative samples play a critical role and are carefully designed. MoCos~\cite{he2020momentum, chen2020improved, chen2021empirical} build a memory bank with a momentum encoder to provide consistent negative samples, yielding promising results for both CNNs~\cite{he2020momentum, chen2020improved} and Vision Transformers~\cite{chen2021empirical}. SimCLR~\cite{chen2020simple} enhances the representation of negative samples with strong data augmentations and a learnable nonlinear projection head. Other methods further combine contrastive learning with instance classification~\cite{cao2020parametric}, data augmentation ~\cite{kalantidis2020hard, zhu2021improving}, clustering~\cite{li2020prototypical,caron2020unsupervised} and adversarial training~\cite{hu2021adco}.

Contrastive learning methods pull positive samples together and push negative samples away, leading to the alignment and uniformity properties of the representation on the hypersphere~\cite{wang2020understanding}. Our work finds that although non-contrastive learning methods are optimized for different objective functions, they share the similar gradient structure with that of contrastive learning.

\vspace{0.5em}\noindent\textbf{Asymmetric Network Methods} aim to accomplish self-supervised learning with only positive pairs~\cite{jean-bastien2020bootstrap,richemond2020byol,chen2021exploring}.
The representational collapse is avoided through the introduction of asymmetric network architecture. BYOL~\cite{jean-bastien2020bootstrap} appends a predictor network after the online branch, and adopts a momentum encoder for the target branch. \cite{richemond2020byol} shows that BYOL is able to achieve competitive performance even without batch statistics. SimSiam~\cite{chen2021exploring} further shows that stopping the gradient to target branch can serve a similar role as the momentum encoder. DINO~\cite{caron2021emerging} adopts an asymmetric pipeline with a self-distillation loss. Despite the impressive performance, little is known about how the asymmetric network can avoid collapse. Recent work~\cite{tian2021understanding} makes a preliminary attempt to analyze the training dynamics, but still lacks a straight-forward explanation.

Based on the conclusion of \cite{tian2021understanding}, our work builds a connection between asymmetric network with contrastive learning methods. From the perspective of backward gradient, we demonstrate the predictor learns to encode the information of previous samples in its weight, which serves as negative gradient during back-propagation. This leads to a similar gradient structure with contrastive learning.

\vspace{0.5em}\noindent\textbf{Feature Decorrelation Methods} are recently proposed for self-supervised learning to prevent the representational collapse~\cite{ermolov2021whitening,zbontar2021barlow,bardes2021vicreg,hua2021feature}. W-MSE~\cite{ermolov2021whitening} whitens feature representations before computing a cosine similarity loss so that the representations are scattered on the unit sphere. Barlow Twins~\cite{zbontar2021barlow} encourages the cross-correlation matrix of representations close to identity matrix, which decorrelates different dimensions of the representation, and strengthens the correlation in the same dimension. VICReg~\cite{bardes2021vicreg} applys variance-invariance-covariance principle to replace the use of batch normalization and cross-correlation matrix. Shuffled-DBN~\cite{hua2021feature} explores the function of batch normalization on the embedding and develops a shuffle method for better feature decorrelation. 

Feature decorrelation methods show comparable results to contrastive learning methods. However, it is still unclear why such approach works well. 
Our work demonstrates that the gradient formulas of feature decorrelation methods can be transformed to a combination of positive and negative samples, and thus share the similar gradient structure with that of contrastive learning.
\section{A Unified Framework for SSL}
\label{sec:method1}

A typical self-supervised learning framework consists of a siamese network. The two branches of the siamese network are named as online branch and target branch, respectively, where target branch representation is served as the training target for the online branch. 
Given the input image $x$, two augmented views $x_1$ and $x_2$ are created as the inputs of the two branches. The encoder $f(\cdot)$ extracts representations $u_i \triangleq f(x_i), i=1, 2$ from these views.

\begin{table}[t]
    \centering
    \resizebox{0.45\textwidth}{!}{
    \begin{tabular}{ll}
    \toprule
        Notation & \multicolumn{1}{c}{Meaning} \\
    \midrule
        $u_1$, $u_2$ & current concerned samples\\
        $v$ & unspecified samples \\
    \midrule
        $u_1^o$, $v^o$ & samples from online branch \\
        $u_2^t$, $v^t$ & samples from unspecified target branch \\
        $u_2^s$, $v^s$ & samples from weight-sharing target branch \\
        $u_2^d$, $v^d$ & samples from stop-gradient target branch \\
        $u_2^m$, $v^m$ & samples from momentum-encoder target branch \\
    \midrule
        $\mathcal{V}$ & unspecified sample set \\
        $\mathcal{V}_{\mathrm{batch}}$ & sample set of current batch \\
        $\mathcal{V}_{\mathrm{bank}}$ & sample set of memory bank \\
        $\mathcal{V}_{\infty}$ & sample set of all previous samples \\
    \bottomrule
    \end{tabular}}
    \caption{Notations used in this paper.}
    \label{tab:notations}
    \vspace{-0.5em}
\end{table}

Table~\ref{tab:notations} illustrates the notations used in this paper. $u_1$ and $u_2$ denote the currently concerned training samples, while $v$ denotes unspecified samples.
$u_1^o$ and $v^o$ denote the representation extracted from the online branch. There are three types of target branches that are widely used: 1) weight-sharing with the online branch, corresponding to $u_2^s$ and $v^s$; 2) weight-sharing but detached from gradient back-propagation, corresponding to $u_2^d$ and $v^d$; 3) momentum encoder updated from the online branch, corresponding to $u_2^m$ and $v^m$. If the target branch type is not specified, $u_2^t$ and $v^t$ are used.
Note that the symmetric loss is always used for the two augmented views as described in \cite{chen2021exploring}.

Moreover, $\mathcal{V}$ represents the sample set considered in current training step. Different methods construct the sample set in different manners: $\mathcal{V}_{\mathrm{batch}}$ contains all samples from current batch, $\mathcal{V}_{\mathrm{bank}}$ consists of a memory bank that stores previous samples, and $\mathcal{V}_{\infty}$ denotes the set of all previous samples, which can be much larger than a memory bank.

Details of gradient analysis can refer to Appendix~\ref{appendix:grad_analysis}.


\subsection{Contrastive Learning Methods}
\label{sec:contrastive}
Contrastive learning methods require negative samples to avoid representational collapse and achieve high performance. They use another view from the same image as the positive sample, and different images as the negative samples. These methods aim to pull positive pairs together while push negative pairs apart. The following InfoNCE loss~\cite{oord2018representation} is usually employed:
\begin{equation}
\label{eq:contrastive_loss}
	L = \mathop{\mathbb{E}}_{u_1,u_2}\bigg[-\log\frac{\exp{(\mathrm{cos}(u_1^o, u_2^t)/\tau)}}{\sum_{v^t\in\mathcal{V}}\exp{(\mathrm{cos}(u_1^o, v^t)/\tau)}}\bigg],
\end{equation}
where the function $\mathrm{cos}(\cdot)$ measures the cosine similarity between two representations, and $\tau$ is the temperature hyper-parameter. Eq.(\ref{eq:contrastive_loss}) can be instantiated for different methods, which we shall discuss below.

\vspace{0.5em}
\noindent\textbf{Relation to MoCo~\cite{he2020momentum, chen2020improved}}. MoCo adopts a momentum encoder for the target branch, and a memory bank to store previous representations from the target branch. Its negative samples come from the memory bank. The gradient for sample $u_1^o$ is therefore:
\begin{equation}
    \label{eq:moco_grad}
	\frac{\partial L}{\partial u_1^o} = \frac{1}{\tau N}\bigg(-u_2^{m} + \sum_{v^{m}\in\mathcal{V}_{\mathrm{bank}}}s_vv^m\bigg),
\end{equation}
where $s_v=\frac{\exp{(\mathrm{cos}(u_1^o, v^m)/\tau)}}{\sum_{y^m\in \mathcal{V}_{\mathrm{bank}}}\exp{(\mathrm{cos}(u_1^o, y^m)/\tau)}}$ is the softmax results over similarities between $u_1^o$ and other samples, and $N$ is the number of all samples in current batch.

\vspace{0.5em}
\noindent\textbf{Relation to SimCLR~\cite{chen2020simple}.} For SimCLR, the target branch shares weights with the online branch, and does not stop the back-propagated gradient. It uses all representations from other images of the same batch as negative samples. Thus, its gradient can be calculated as:
\begin{equation}
\label{eq:simclr_grad}
\begin{split}
    \frac{\partial L}{\partial u_1^o} = &\frac{1}{\tau N}\bigg(-u_2^{s} + \sum_{v^s\in \mathcal{V}_{\mathrm{batch}}\setminus u_1^o}s_vv^s\bigg) \\
    &+ \color{blue}{\underbrace{\color{black}{\frac{1}{\tau N}\bigg(-u_2^{s} + \sum_{v^s\in \mathcal{V}_{\mathrm{batch}}\setminus u_1^o}t_vv^s\bigg)}}_{\text{reduce to }0}}\color{black}{,}
\end{split}
\end{equation}
where $t_v=\frac{\exp{(\mathrm{cos}(v^s, u_1^o)/\tau)}}{\sum_{y^s\in \mathcal{V}_{\mathrm{batch}}\setminus v^s}\exp{(\mathrm{cos}(v^s, y^s)/\tau)}}$ is computed over similarities between sample $v^s$ and its contrastive samples $\mathcal{V}_{\mathrm{batch}}\setminus v^s$. If the gradient through the target branch is stopped, the second term in Eq.(\ref{eq:simclr_grad}) will vanish. We have verified that stopping the second gradient term will not affect the performance (see Appendix Table~\ref{tab:grad_sim}), so Eq.(\ref{eq:simclr_grad}) can be simplified to only the first term.

\vspace{0.5em}
\noindent\textbf{Unified Gradient.} From the perspective of gradient, above methods can be represented in a unified form:
\begin{equation}
    \label{eq:contrastive_grad}
	\frac{\partial L}{\partial u_1^o} = \frac{1}{\tau N}\bigg(-u_2^t + \sum_{v^t\in \mathcal{V}}s_vv^t\bigg),
\end{equation}
where the gradient is made up of a weighted sum of positive and negative samples. The effect of $-u_2^t$ is to pull positive samples together, and the effect of $\sum_{v^t\in \mathcal{V}}s_vv^t$ is to push negative samples apart. We name these two terms as the positive and negative gradient, respectively. The only difference between methods is what type of target branch is used and how the contrastive sample set $\mathcal{V}$ is built.


\subsection{Asymmetric Network Methods} 
Asymmetric network methods learn powerful representations by maximizing the similarity of positive pairs, without using negative samples. Such methods need symmetry-breaking network designs to avoid representational collapse. To achieve this, a predictor $h(\cdot)$ is appended after the online branch. The gradient to the target branch is also stopped. The objective function can be presented as:
\begin{equation}
    \label{eq:asymmetric_loss}
	L = \mathop{\mathbb{E}}_{u_1,u_2}\bigg[-\mathrm{cos}(h(u_1^o), u_2^t)\bigg].
\end{equation}

\vspace{0.5em}
\noindent\textbf{Relation to BYOL~\cite{jean-bastien2020bootstrap}.} For BYOL, a momentum encoder is used for the target branch, \ie, $u_2^t=u_2^m$ in Eq.(\ref{eq:asymmetric_loss}).

\vspace{0.5em}
\noindent\textbf{Relation to Simsiam~\cite{chen2021exploring}.} Simsiam shows that momentum encoder is not necessary, and only applies the stop-gradient operation to the target branch, \ie, $u_2^t=u_2^d$ in Eq.(\ref{eq:asymmetric_loss}).

\vspace{0.5em}
\noindent\textbf{Unified Gradient.} While asymmetric network methods have achieved impressive performance, it is unclear how these methods avoid collapse solution. Recently, DirectPred~\cite{tian2021understanding} makes a preliminary attempt towards this goal via studying the training dynamics. It further proposes an analytical solution for the predictor $h(\cdot)$.

Specifically, DirectPred claims that the predictor can be formulated as $h(v)=W_h v$, where  $W_h$ can be directly calculated based on the correlation matrix $\mathbb{E}_v(v v^T)$. In practice, this correlation matrix is calculated as the moving average of the correlation matrix for each batch, \ie, $F\triangleq \sum_{v^o \in \mathcal{V}_{\infty}} \rho_vv^o {v^o}^T$, where $\rho_v$ is the moving average weight for each sample according to their batch order. By decomposing $F$ into its eigenvalues $\Lambda_F$ and eigenvectors $U$, $W_h$ can be calculated as 
\begin{equation}
    \label{eq:directpred}
    W_h = U\Lambda_h U^T, \ \ \Lambda_h = \Lambda_{F}^{1/2}+\epsilon \lambda_{max}I,
\end{equation}
where $\lambda_{max}$ is the max eigenvalue of $F$ and $\epsilon$ is a hyper-parameter to help boost small eigenvalues.

While DirectPred shows what the predictor learns, We step further and try to reveal the relationship between the predictor and contrastive learning. With the help of DirecPred, the gradient can be derived and simplified as:
\begin{equation}
\small
    \label{eq:asymmetric_grad_simplified}
	\frac{\partial L}{\partial u_1^o} = \frac{1}{||W_h u_1^o||_2N}\bigg(-W_h^T u_2^t+\lambda\sum_{v^o \in \mathcal{V}_{\infty}} (\rho_v{u_1^o}^Tv^o)v^o\bigg),
\end{equation}
where $-W_h^T u_2^t$ and $\sum_{v^o \in \mathcal{V}_{\infty}} (\rho_v{u_1^o}^Tv^o)v^o$ work as the positive and negative gradient respectively and  $\lambda=\frac{{u_1^o}^T W_h^T u_2^t}{{u_1^o}^T(F + \epsilon^2I)u_1^o}$ is a balance factor.

It seems counter-intuitive that Eq.(\ref{eq:asymmetric_grad_simplified}) is also a combination of positive and negative samples, since no negative samples appear in the loss function explicitly. In fact, they come from the optimization of the predictor network. From the findings of \cite{tian2021understanding}, the eigenspace of the predictor $W_h$ will gradually align with that of the feature correlation matrix $F$. Hence the predictor may learn to encode the information of correlation matrix in its parameters. During back-propagation, the encoded information will work as negative gradient and contribute to the direction of optimization.


\subsection{Feature Decorrelation Methods}
Feature decorrelation methods emerge recently as a new solution to self-supervised learning. It proposes to reduce the redundancy among different feature dimensions so as to avoid collapse. Recent works adopt different loss forms for feature decorrelation. We discuss their relations below.

\vspace{0.5em}
\noindent\textbf{Relation to Barlow Twins~\cite{zbontar2021barlow}}.
Barlow Twins utilizes the following loss function:
\begin{equation}
\label{eq:bt_loss}
	L = \sum_{i=1}^C{(W_{ii}-1)^2+\lambda\sum_{i=1}^C{\sum_{j\ne{i}}}{W_{ij}^2}},
\end{equation}
where $W=\frac{1}{N}\sum_{v_1^o, v_2^s\in\mathcal{V}_{\mathrm{batch}}}v_1^ov_2^{sT}$ is a cross-correlation matrix, $C$ denotes the number of feature dimensions and $\lambda$ is a balancing hyper-parameter. The diagonal elements of $W$ are encouraged to be close to $1$, while those off-diagonal elements are forced to be close to $0$.

At first glance, Eq.(\ref{eq:bt_loss}) is drastically different from loss functions of previous methods. However, it actually works in the same way from the view of gradient, which can be calculated as
\begin{equation}
\label{eq:bt_grad}
	\frac{\partial L}{\partial u_1^o} =  \frac{2}{N} \bigg(\color{blue}{\underbrace{\color{black}{-Au_2^s}}_{\text{reduce to }-0.1u_2^s}}\color{black}+\lambda\sum_{v_1^o, v_2^s\in\mathcal{V}_{\mathrm{batch}}}{\frac{u_2^{sT}v_2^s}{N}v_1^o}\bigg),
\end{equation}
where $A = I-(1-\lambda)W_{\mathrm{diag}}$. Here $(W_{\mathrm{diag}})_{ij} = \delta_{ij} W_{ij}$ is the diagonal matrix of $W$, where $\delta_{ij}$ is the Kronecker delta.

We plot the max and min values of $W_{\mathrm{diag}}$ in Figure~\ref{fig:justification}(c), which shows $W_{\mathrm{diag}}$ is close to a scaled identity matrix. Therefore, we replace $A$ with an identity matrix multiplied by $0.1$ in practice. It has been verified that such replacement actually does no harm to the final result (see Table~\ref{tab:methods_comparison}(g)).
In addition, it should be noted that Barlow Twins applies batch normalization rather than $\ell_2$ normalization to the representation $v$. We have verified that changing to $\ell_2$ normalization will not affect the performance (see Table~\ref{tab:methods_comparison}(h)).

\vspace{0.5em}
\noindent\textbf{Relation to VICReg~\cite{bardes2021vicreg}}.
VICReg does a few modifications to Barlow Twins with the following loss function:
\begin{equation}
\label{eq:vic_loss}
\vspace{-0.5em}
\begin{split}
    L = &\frac{1}{N}\sum_{v_1^o, v_2^s\in\mathcal{V}_{\mathrm{batch}}}||v_1^o-v_2^s||_2^2 + \frac{\lambda_1}{c}\sum_{i=1}^c\sum_{j\ne i}^cW_{ij}'^2\\
    &+ \frac{\lambda_2}{c}\sum_{i=1}^c\max(0, \gamma - \mathrm{std}(v_1^o)_i),
\end{split}
\vspace{-0.5em}
\end{equation}
where $W'=\frac{1}{N-1}\sum_{v_1^o\in\mathcal{V}_{\mathrm{batch}}}(v_1^o-\bar{v}_1^o)(v_1^o-\bar{v}_1^o)^{T}$ is the covariance matrix of the same view, $\mathrm{std}(v)_i$ denotes the standard deviation of the $i$-th channel of $v$, $\gamma$ is a constant target value for it, and $\lambda_1$, $\lambda_2$ are balancing weights.

Similarly, its gradient can be derived as:
\begin{equation}
\label{eq:vic_grad}
\vspace{-0.5em}
\begin{split}
    \frac{\partial L}{\partial u_1^o} = &\frac{2}{N}\bigg(-u_2^s + \lambda\sum_{v_1^o\in\mathcal{V}_{\mathrm{batch}}}\frac{\tilde{u}_1^{oT}\tilde{v}_1^o}{N}\tilde{v}_1^o\bigg)\\
    & +\color{blue}\underbrace{\color{black}{\frac{2\lambda}{N}\bigg(\frac{1}{\lambda}u_1^o - B\tilde{u}_1^o\bigg)}}_{\text{reduce to }0}\color{black},
\end{split}
\vspace{-0.5em}
\end{equation}
where $\tilde{v} = v-\bar{v}$ is the de-centered sample, $\lambda=\frac{2\lambda_1N^2}{c(N-1)^2}$, and $B=\frac{N}{c\lambda(N-1)}(2\lambda_1W'_{\mathrm{diag}} + \frac{\lambda_2}{2}\mathrm{diag}(\mathbbm{1}(\gamma-\mathrm{std}(v_1^o)>0)\oslash\mathrm{std}(v_1^o)))$. Here $\mathrm{diag}(x)$ is a matrix with diagonal filled with the vector $x$, $\mathbbm{1}(\cdot)$ is the indicator function, and $\oslash$ denotes element-wise division.

VICReg does not apply any normalization on $v$, and instead requires the de-center operation and standard deviation term in the loss function. We have verified that it is able to get rid of such terms by employing $\ell_2$ normalization on $v$ (see Table~\ref{tab:methods_comparison}(j)). In fact, we have plotted the cosine similarity between the neglected term and $\sum_{v_1^o\in\mathcal{V}_{\mathrm{batch}}}\frac{\tilde{u}_1^{oT}\tilde{v}_1^o}{N}\tilde{v}_1^o$ in Figure~\ref{fig:justification}(d). They are expected to have similar effect on the training because of similar direction. Eq.(\ref{eq:vic_grad}) can be reduced to only the first term without de-center operation.

\vspace{0.5em}
\noindent\textbf{Unified Gradient.} Because $v^s$ and $v^o$ are mathematically equivalent, the gradient form of feature decorrelation family can be unified as:
\begin{equation}
    \frac{\partial L}{\partial u_1^o} = \frac{2}{N}\bigg(-u_2^t+\lambda\sum_{v^o\in\mathcal{V}_{\mathrm{batch}}}\frac{u^{oT}v^o}{N}v^o_1\bigg),
\end{equation}
where the first term $-u_2^t$ acts as the positive gradient, the second term $\sum_{v^o\in\mathcal{V}_{\mathrm{batch}}}{\big(u^{oT}v^o/N\big)v^o_1}$ is the negative gradient, and $\lambda$ is also a balance factor. The only difference between methods is the subscript for negative coefficient.
Feature decorrelation methods actually work in a similar way with other self-supervised methods. The positive and negative gradient come from the diagonal and off-diagonal elements of the correlation matrix.
\section{Key Factors in SSL}
\label{sec:method2}
As we analyzed before, the gradients for different self-supervised learning methods share a similar formula:
\begin{equation}
    \label{eq:unified-gradient}
    \frac{\partial L}{\partial u_1^o} = \nabla L_p + \lambda \nabla L_n,
\end{equation}
where the gradient consists of three components: the positive gradient $\nabla L_p$, the negative gradient $\nabla L_n$ and the balance factor $\lambda$. However, there are still differences on the specific form of these three components, and a natural question arises: will the gradient form affect the performance of self-supervised learning? Furthermore, although these methods share similar gradient formula, they usually differ from each other on the type of target branch and the construction of sample set $\mathcal{V}$. In this section, we shall conduct a thorough comparison between these methods and present the key factors that influence the final performance.

Although previous works have compared their methods with others, the training settings are usually different. To provide a fair comparison, we use a unified training and evaluation setting, in which only the loss function is changed. Our setting mainly follows \cite{chen2021exploring} (see Appendix~\ref{appendix:implementation}).

\begin{table*}[t]
  \centering
  \resizebox{0.99\textwidth}{!}{
  \begin{tabular}{@{}cccccccc@{}}
    \toprule
    & Method & Norm & Pos Grad & Balance Factor & Neg Grad & Sample Set & Linear Eval \\
    \midrule
    \multicolumn{8}{l}{Contrastive learning methods} \\
    \midrule
    (a) & MoCo~\cite{he2020momentum} & $\ell_2$ & $-u_2^m$ & 1 & $\sum_{v^m\in\mathcal{V}_{\mathrm{bank}}}s_vv^m$ & $\mathcal{V}_{\mathrm{bank}}$ & 70.0 \\
    (b) & SimCLR$^*$~\cite{chen2020simple} & $\ell_2$ & $-u_2^m$ & 1 & $\sum_{v^m\in\mathcal{V}_{\mathrm{batch}}\setminus u_1^o}s_vv^m$ & $\mathcal{V}_{\mathrm{batch}}\setminus u_1^o$ &  70.0 \\
    \midrule
    \multicolumn{8}{l}{Asymmetric network methods} \\
    \midrule
    (c) & BYOL~\cite{jean-bastien2020bootstrap} & $\ell_2$ & -   & - & - & - & 70.3 \\
    (d) & BYOL(DirectPred~\cite{tian2021understanding}) & $\ell_2$ & $-W_h^Tu_2^m$   & $\frac{u_1^{oT} W_h^T u_2^m}{u_1^{oT}(F+\epsilon^2I)u_1^o}$ & $\sum_{v^o\in\mathcal{V}_{\infty}}(\rho_vu_1^{oT}v^o)v^o$ & $\mathcal{V}_{\infty}$ & 70.2 \\
    (e) & - & $\ell_2$ & $-u_2^m$   & 100 & $\sum_{v^o\in\mathcal{V}_{\infty}}(\rho_vu_1^{oT}v^o)v^o$ & $\mathcal{V}_{\infty}$ & 70.3 \\
    \midrule
    \multicolumn{8}{l}{Feature decorrelation methods} \\
    \midrule
    (f) & Barlow Twins$^*$~\cite{zbontar2021barlow} & BN & $-Au_2^m$ & 5$\times$10$^{-3}$ & $\sum_{v_1^o, v_2^m\in\mathcal{V}_{\mathrm{batch}}}{\frac{u_2^{mT}v_2^m}{N}v_1^o}$ & $\mathcal{V}_{\mathrm{batch}}$ & 69.0 \\
    (g) & - & BN & $-u_2^m$ & 5$\times$10$^{-3}$ & $\sum_{v_1^o, v_2^m\in\mathcal{V}_{\mathrm{batch}}}{\frac{u_2^{mT}v_2^m}{N}v_1^o}$ & $\mathcal{V}_{\mathrm{batch}}$ & 69.7 \\
    (h) & - & $\ell_2$ & $-u_2^m$ & 50 & $\sum_{v_1^o, v_2^m\in\mathcal{V}_{\mathrm{batch}}}{\frac{u_2^{mT}v_2^m}{N}v_1^o}$ & $\mathcal{V}_{\mathrm{batch}}$ & 70.0 \\
    (i) & VICReg$^*$~\cite{bardes2021vicreg} & - & $-u_2^m$ & 4$\times$10$^{-\mathrm{5}}$ & $\sum_{v_1^o\in\mathcal{V}_{\mathrm{batch}}}{\frac{u_1^{oT}v_1^o}{N}v_1^o} + \frac{1}{\lambda}u_1^o - B\tilde{u}_1^o$ & $\mathcal{V}_{\mathrm{batch}}$ & 70.0 \\
    (j) & - & $\ell_2$ & $-u_2^m$ & 25 & $\sum_{v_1^o\in\mathcal{V}_{\mathrm{batch}}}{\frac{u_1^{oT}v_1^o}{N}v_1^o}$ & $\mathcal{V}_{\mathrm{batch}}$ & 69.8 \\
    \bottomrule
  \end{tabular}}
  \caption{Performance comparison for different methods on ImageNet~\cite{deng2009imagenet}. ``Norm'' denotes the normalization applied to the representations before loss calculation. ``Pos Grad'', ``Balance Factor'' and ``Neg Grad'' correspond to the components in Eq.~(\ref{eq:unified-gradient}). Linear evaluation follows the protocol in \cite{chen2021exploring}. $^*$Note that momentum encoder is used for target branch.}
  \label{tab:methods_comparison}
  \vspace{-1.0em}
\end{table*}

\subsection{Gradient Form}
\label{sec:gradient_exp}
We first explore how much difference the gradient form can make in different methods. For a fair comparison, the target branch adopts momentum encoder for all methods. The effect of target branch type will be discussed in Section~\ref{sec:target_exp}.
It should be noted that the we apply momentum encoder in the loss form and derive the corresponding gradients, so some negative gradient forms do not contain $v^m$. 

Specifically, we first try to compare and simplify the gradient form within each type of method. This can filter out irrelevant elements at early stage and make the comparison more clear. After that we can compare these methods all together. Because the scales of positive and negative gradients can vary a lot during simplification, we search for the best balance factor for each combination. 

\vspace{0.3em}
\noindent\textbf{Simplification for Contrastive Learning.} 
Table~\ref{tab:methods_comparison}(ab) report the performance of different contrastive learning methods. Original MoCo~\cite{he2020momentum} is used in Table~\ref{tab:methods_comparison}(a). Because momentum encoder is applied to SimCLR~\cite{chen2020simple} in Table~\ref{tab:methods_comparison}(b), the second term in Eq.(\ref{eq:simclr_grad}) naturally diminishes. These two methods show nearly no differences on the final results.

We also note that SimCLR uses $\mathcal{V}_{\mathrm{batch}}$ rather than $\mathcal{V}_{\mathrm{bank}}$ as in MoCo, but there is only minor difference. This suggests that with proper training setting, larger number of negative samples may not be necessary for good performance.

\vspace{0.3em}
\noindent\textbf{Simplification for Asymmetric Network.} 
Table~\ref{tab:methods_comparison}(c-e) give the simplification results for asymmetric network methods. The original BYOL~\cite{jean-bastien2020bootstrap} and the gradient version of BYOL with DirectPred~\cite{tian2021understanding} form are presented in Table~\ref{tab:methods_comparison}(cd), respectively, whose results are consistent with the conclusion of \cite{tian2021understanding}. SimSiam~\cite{chen2021exploring} is not presented here, because its momentum encoder variant is just BYOL.

In Table~\ref{tab:methods_comparison}(e) we substitute $W_h$ in the positive gradient with identity matrix, and reduce the dynamic balance factor to a constant scalar. Such replacement does not lead to performance degradation. Therefore, the gradient form of asymmetric network methods can be unified as Table~\ref{tab:methods_comparison}(e).

\vspace{0.3em}
\noindent\textbf{Simplification for Feature Decorrelation.} 
We demonstrate the results of feature decorrelation methods in Table~\ref{tab:methods_comparison}(f-j). For Barlow Twins~\cite{zbontar2021barlow}, the matrix $A$ in the positive gradient of Table~\ref{tab:methods_comparison}(f) is first substituted with identity matrix in Table~\ref{tab:methods_comparison}(g). 
The results imply that this will not harm the performance. In Table~\ref{tab:methods_comparison}(h),
batch normalization is then replaced with $\ell_2$ normalization, and no accuracy decrease is observed.

For VICReg~\cite{bardes2021vicreg}, we report its result in Table~\ref{tab:methods_comparison}(i). In Table~\ref{tab:methods_comparison}(j), $\ell_2$ normalization is applied to the representation, and the $\lambda_1u_1-B\tilde{u}_1$ term is removed from negative gradient. Such simplification produces similar result.

In the end, Table~\ref{tab:methods_comparison}(hj) only differ in how to calculate negative coefficients. The comparison indicates that similar performances can be obtained. Thus, the gradient form of feature decorrelation methods can be unified as Table~\ref{tab:methods_comparison}(j).

\vspace{0.3em}
\noindent\textbf{Comparison between Different Methods.} 
Finally, we can compare different kinds of methods with their unified gradient form, \ie, Table~\ref{tab:methods_comparison}(bej). Among three components of gradient, they share the same positive gradient, the balance factor is searched for the best one, and the only difference is the negative gradient. Table~\ref{tab:methods_comparison} shows that the performance gap between different methods is actually minor ($<$0.5$\%$ points). What's more, asymmetric network methods are similar with feature decorrelation methods on gradient form, but utilize $\mathcal{V}_{\infty}$ instead of $\mathcal{V}_{\mathrm{batch}}$. This implies the construction of $\mathcal{V}$ is not vital for self-supervised learning.

\subsection{Target Branch Type}
\label{sec:target_exp}
The type of target branch is distinct for different methods in their original implementation. 
In Section~\ref{sec:gradient_exp}, we adopts momentum encoder for all methods. Now, we study the effect of different target branch types in Table~\ref{tab:effect_of_encoder}. There can be three choices for the target branch: weight-sharing, stop-gradient and momentum-encoder. We use the unified form (\ie, Table~\ref{tab:methods_comparison}(bej)) as representatives for these three kinds of methods, and change the target branch type. 
Because a symmetric loss is always employed, the weight-sharing and stop-gradient variants of the gradient form are actually the same. We omit the weight-sharing variant for simplicity.

For the stop-gradient target branch type, the results for different self-supervised learning methods are very similar, which is consistent with the conclusion in Section~\ref{sec:gradient_exp}. 
For the momentum-encoder target branch type, it can improve the performance of all three kinds of methods with $\sim 2\%$ points compared to the stop-gradient target branch type. This shows that momentum encoder is beneficial for these self-supervised learning methods, and can provide a consistent performance gain.

We further consider which part of gradient the momentum encoder has effect on. To achieve this, we only adopt momentum encoder output for the positive gradient. Table~\ref{tab:effect_of_encoder} indicates that it's enough to apply momentum encoder to the positive gradient. This suggests that a consistent and slow-updating positive goal may be very important for self-supervised learning.

\begin{table}[t]
    \centering
    \resizebox{0.48\textwidth}{!}{
    \begin{tabular}{@{}ccccc@{}}
    \toprule
    \multirow{2}{*}{Pos Grad} &  \multirow{2}{*}{Neg Grad}  & \makecell{Contrastive\\ Learning} & \makecell{Asymmetric\\Network} & \makecell{Feature\\Decorrelation} \\
     &   & Table~\ref{tab:methods_comparison}(b) & Table~\ref{tab:methods_comparison}(e) & Table~\ref{tab:methods_comparison}(j) \\
    \midrule
    \multicolumn{2}{c}{stop gradient} & 67.6 & 67.9    & 67.6 \\ 
    \multicolumn{2}{c}{momentum} & 70.0 & 70.2 & 69.8 \\
    momentum & stop gradient & 70.1  & 70.3 & 69.8 \\
    \bottomrule
    \end{tabular}}
    \caption{Effect of target branch type. We report ImageNet~\cite{deng2009imagenet} linear evaluation accuracy after 100-epoch pre-training.}
    \label{tab:effect_of_encoder}
    \vspace{-0.8em}
\end{table}
\section{A Concise Gradient Form for SSL}
\label{sec:exp}
\subsection{UniGrad}
The comparison between gradients of different methods leads us to find a concise but effective gradient form for self-supervised learning. The  proposed gradient, named {\name}, can be represented as  
\begin{equation}
\small
    \label{eq:\name-gradient}
    \frac{\partial L}{\partial u_1^o} = -u_2^m + \lambda Fu_1^o,
\end{equation}
where $F= \sum_{v^o \in \mathcal{V}_{\infty}} \rho_vv^o v^{oT}$. Note that this gradient form is exactly the one described in Table~\ref{tab:methods_comparison}(e).

To fully understand this gradient, we give analysis through its corresponding object function:
\begin{equation}
\small
\label{eq:\name-loss}
    L = \mathop{\mathbb{E}}_{u_1,u_2}\bigg[-\mathrm{cos}(u_1^o,u_2^m)+\frac{\lambda}{2}\sum_{v^o \in \mathcal{V}_{\infty}} \rho_v \mathrm{cos}^2(u_1^o,v^o) \bigg],
\end{equation}
where $\lambda$ is set to 100 as default.
The objective function consists of two terms. The first term maximizes the cosine similarity between positive samples, which encourages modeling the invariance with respect to data augmentations. The second term expects the similariry between negative samples close to zero so as to avoid representational collapse.

\vspace{0.5em}
\noindent\textbf{Relation to Contrastive Learning.} Compare to the InfoNCE~\cite{oord2018representation} used in MoCo~\cite{he2020momentum,chen2020improved} and SimCLR~\cite{chen2020simple},
{\name} expects the similarity with negative samples close to zero to avoid collapse, while the InfoNCE encourages the similarity with negative samples to be lower than that with positive samples as much as possible. Moreover, {\name} could encode infinite negative samples via a correlation matrix with less memory cost compared to a memory bank. 

\vspace{0.5em}
\noindent\textbf{Relation to Asymmetric Network.} 
Compare to BYOL~\cite{jean-bastien2020bootstrap} and SimSiam~\cite{chen2021exploring}, our method could learn meaningful representations without the need of a predictor, thus gets rid of additional optimization tricks (usually a larger learning rate is needed for the predictor) and potential influence of the design of this predictor. Compare to DirectPred~\cite{tian2021understanding} with a optimization-free predictor, {\name} removes the need for SVD decomposition.

\begin{figure*}[t]
\begin{subfigure}{.24\textwidth}
    \centering
    \includegraphics[width=1.0\textwidth]{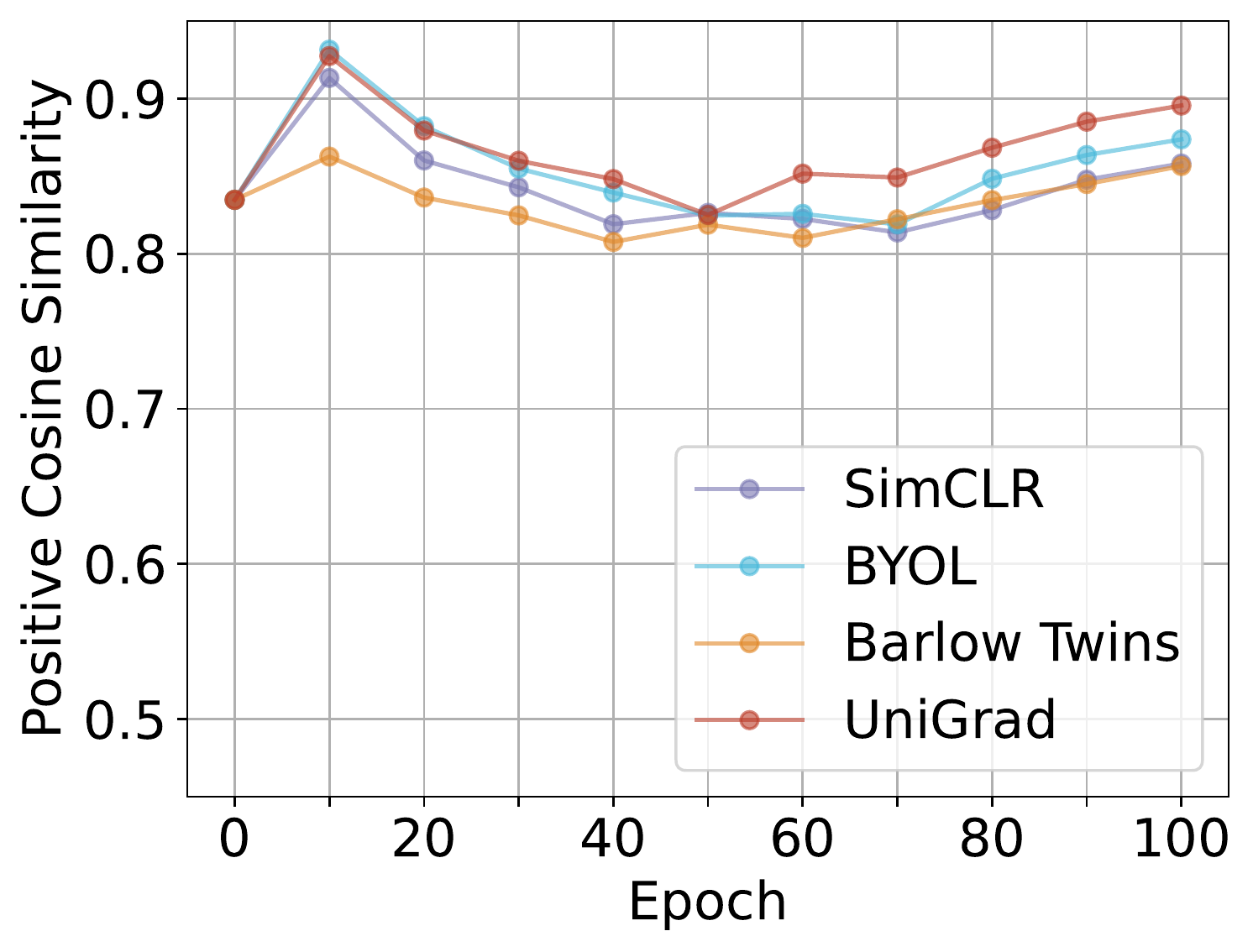}
    \subcaption{positive cosine similarity}
\end{subfigure}
\begin{subfigure}{.24\textwidth}
    \centering
    \includegraphics[width=1.0\textwidth]{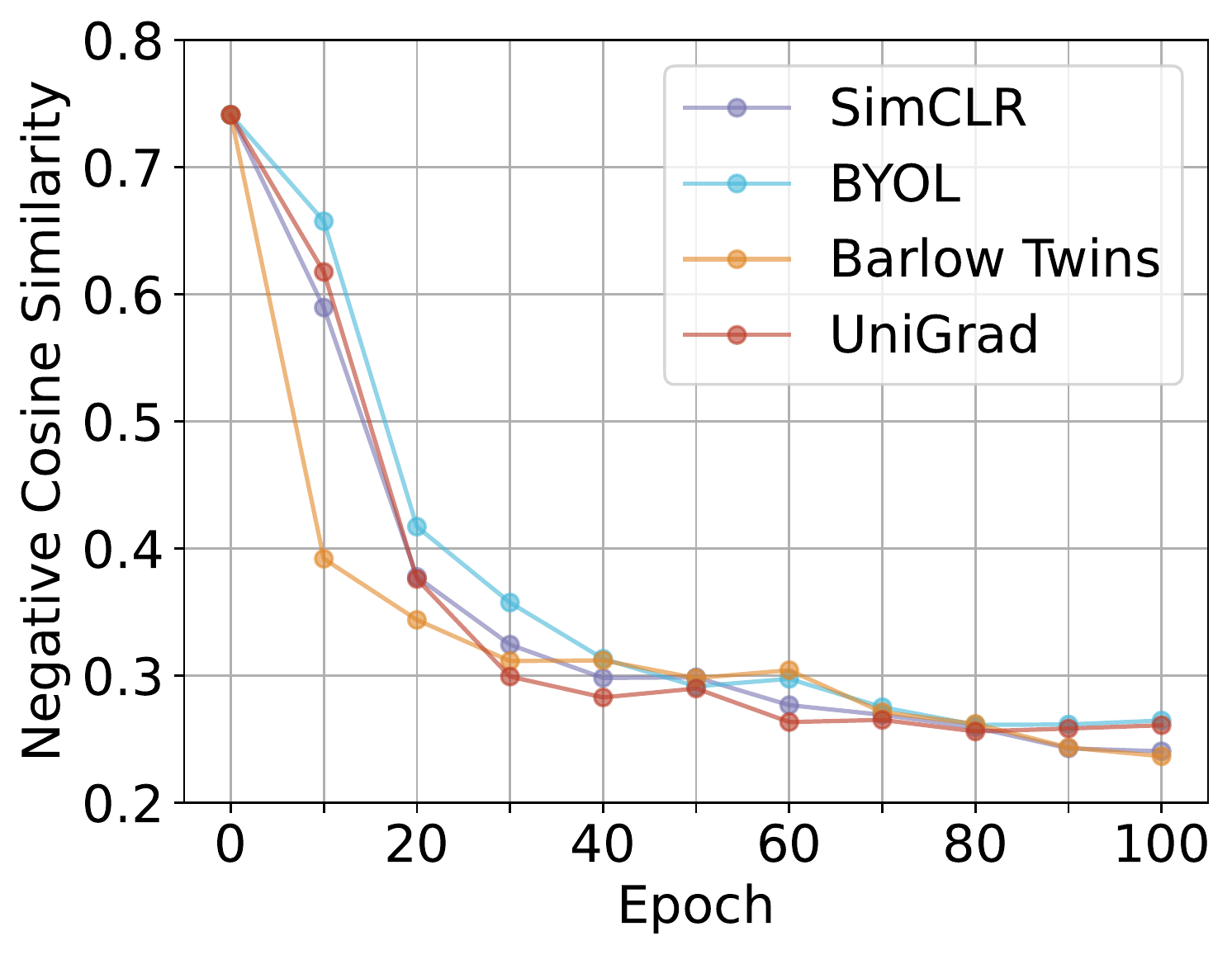}
    \subcaption{negative cosine similarity}
\end{subfigure}
\begin{subfigure}{.24\textwidth}
    \centering
    \includegraphics[width=1.0\textwidth]{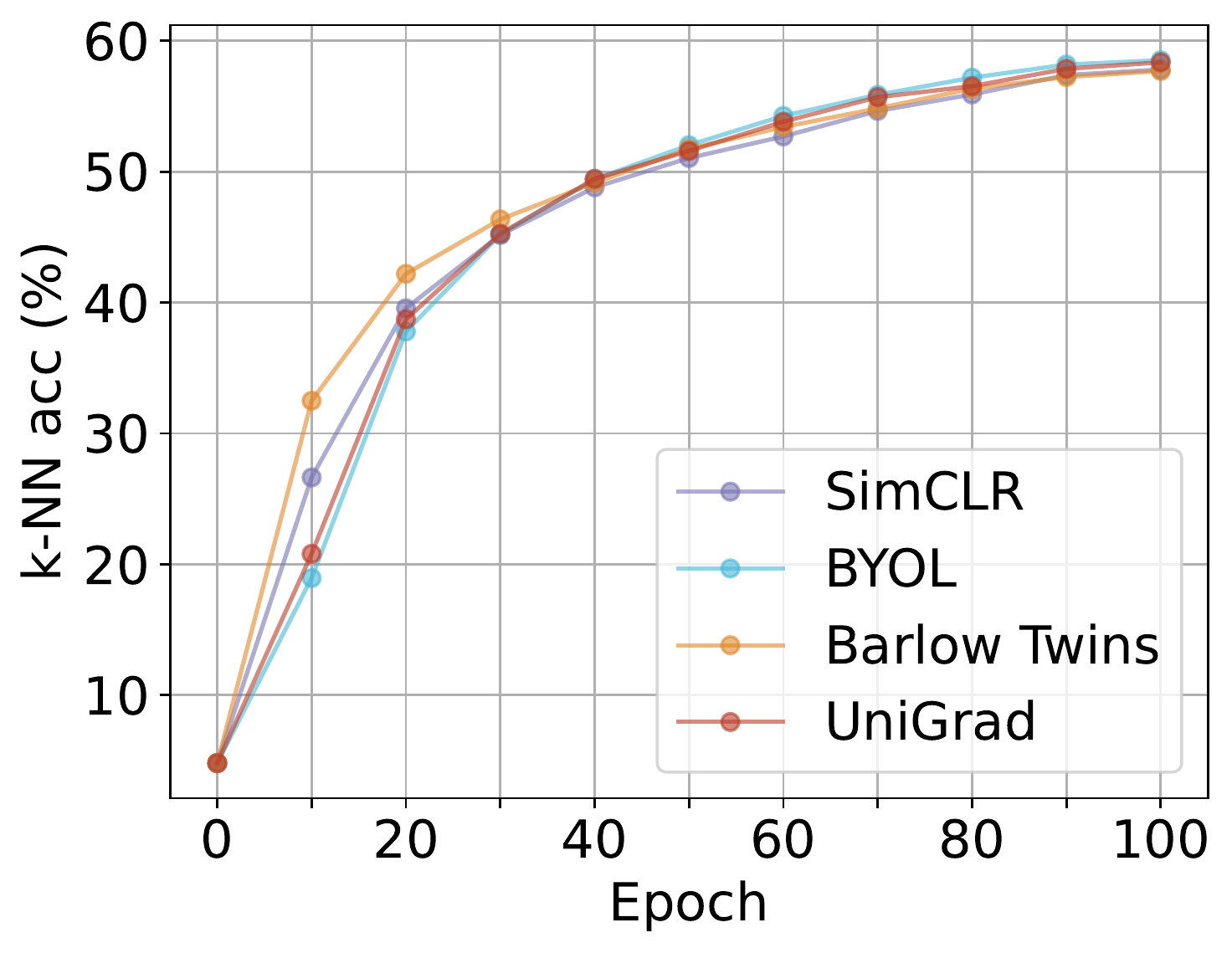}
    \subcaption{k-NN accuracy}
\end{subfigure}
\begin{subfigure}{.24\textwidth}
    \centering
    \includegraphics[width=1.0\textwidth]{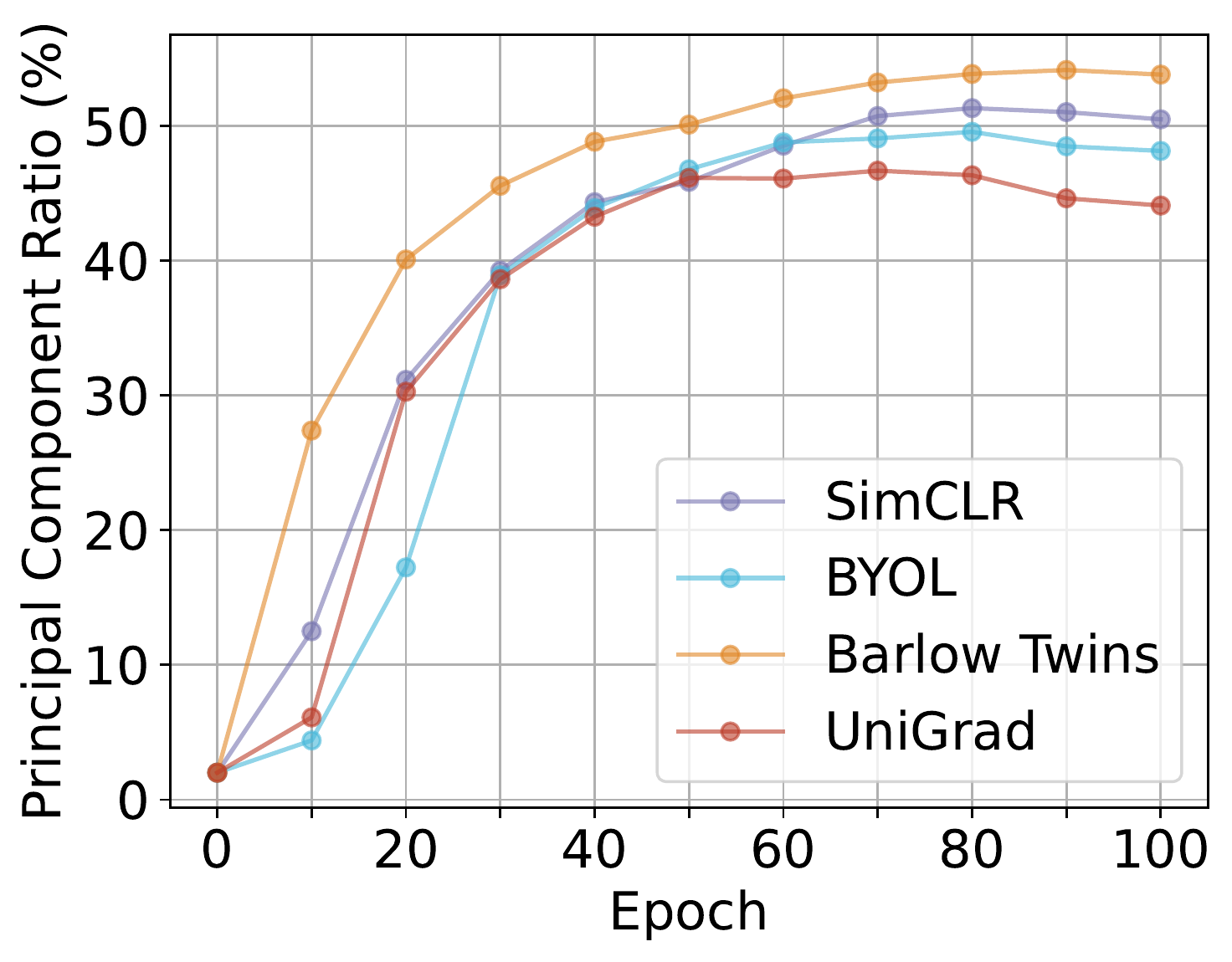}
    \subcaption{principal component ratio}
\end{subfigure}
    \vspace{-0.5em}
    \caption{Learning trajectory for different methods. The metric of principal component ratio is to evaluate feature decorrelation degree. We apply PCA to representations, and count the number of eigenvalues whose cumulative sum first exceeds $90\%$.}
    \label{fig:cos_sim}
    \vspace{-0.8em}
\end{figure*}

\vspace{0.5em}
\noindent\textbf{Relation to Feature Decorrelation.} Compare to Barlow Twins~\cite{zbontar2021barlow} and VICReg~\cite{bardes2021vicreg}, {\name} could achieve a similar effect to decorrelate different channels without direct optimization of the covariance or cross-correlation matrix (see Figure~\ref{fig:cos_sim}). In addition, our method uses $\ell_2$ normalization instead of batch normalization or extra restrictions on the variance of each channel.

\vspace{0.5em}
\noindent\textbf{Discussion.} Since we have observed close performance achieved by {\name} and other methods in Table~\ref{tab:methods_comparison}, we wonder if the representations learned via various losses could end up with similar properties. In Figure~\ref{fig:cos_sim}, We compare the learning trajectory of different methods from the aspects of the similarity between positive/negative pairs, the k-NN accuracy and the degree of feature decorrelation. We find that there is no significant difference between {\name} and other methods. The result implies that these methods work in a similar mechanism, which coincides with the comparison of their gradients in Section~\ref{sec:method2}. For instance, SimCLR and BYOL can also learn to decorrelate different channels and Barlow Twins can learn to discriminate positive and negative samples as well. Besides its competitive performance, our method works as a concise version connected to these three kinds of methods without complicated components, such as memory bank and predictor.

\subsection{Application on Data Augmentations}
Benefiting from its concise form, {\name} can be easily extended with commonly used data augmentations~\cite{zhang2018mixup,yun2019cutmix,li2021unsupervised,wang2020understanding,caron2020unsupervised,caron2021emerging} to further boost its performance. As a demonstration, we show how to apply CutMix~\cite{yun2019cutmix,li2021unsupervised} and multi-crop~\cite{caron2020unsupervised,caron2021emerging} to our method below. 

\textit{CutMix} generates new samples by replacing a randomly selected image region with a patch from another image. 
Given a batch of images, we cut patches from this batch with shuffled order, and paste them onto the original batch. For these mixed images, their positive gradients are calculated from the normal images and then mixed according to the mixup ratio. $F$ is calculated from normal images only.

\textit{Multi-crop} samples additional smaller-sized crops to increase the number of views of an image. Specifically, we use $2\times224$ global views and $6\times96$ local views, with global scale set to $(0.4, 1)$ and local scale set to $(0.05, 0.4)$ respectively.
For each global view, its positive gradient comes from the other global view. 
For each local view, its positive samples consist of the average of two global views. $F$ is calculated from global views only.

\begin{table}
    \centering
    \resizebox{0.46\textwidth}{!}{
    \begin{tabular}{@{}lccc@{}}
    \toprule
    {Method} &  Epoch  & Time & Linear Eval \\
    \midrule
    \name & 100  & 38.2h & 70.3 \\
    \name+CutMix & 100 & 38.2h & 71.2\\
    \name+multi-crop & 100 & 114.6h & 71.7\\
    \name+CutMix+multi-crop & 100 & 114.6h & 72.3\\
    \bottomrule
    \end{tabular}}
    \vspace{-0.5em}
    \caption{Ablation on CutMix and multi-crop.}
    \label{tab:ablation}
    \vspace{-0.5em}
\end{table}

\vspace{0.3em}
\noindent\textbf{Ablation Study.} We first conduct ablation study to validate the impact of CutMix and multi-crop on {\name} under the experiments setting described in Section~\ref{sec:method2}. As shown in Table~\ref{tab:ablation}, CutMix and multi-crop achieve an improvement of 0.9\% and 1.4\% respectively, and combining these two strategies together boosts the improvement to 2.0\%. 
We also report the training time in Table~\ref{tab:ablation}. The implementation of CutMix only adds negligible training overhead compared to normal training, while multi-crop introduces a relatively heavy training cost. These variants can be used according to the available computational resources.

\vspace{0.3em}
\noindent\textbf{More Training Epoches.} 
We evaluate the performance of our method with more training epoches. We adopt another set of training setting for faster pretraining (see Appendix~\ref{appendix:implementation}). The linear evaluation setting follows Section~\ref{sec:method2}. Table~\ref{tab:sota} compares our results with previous methods. \name\ with CutMix can already surpass other methods that do not use multi-crop. By further employing multi-crop, it shows comparable performance with current state-of-the-art methods. We also transfer the pre-trained model to downstream tasks, including semi-supervised learning on ImageNet~\cite{deng2009imagenet} and object detection on PASCAL VOC~\cite{everingham2010pascal} and COCO~\cite{lin2014microsoft}. Our model is able to achieve competitive results with other leading methods (see Appendix~\ref{appendix:results}).

\begin{table}
    \centering
     \resizebox{0.4\textwidth}{!}{
    \begin{tabular}{@{}lccc@{}}
    \toprule
    {Method} &  Epoch  & Linear Eval \\
    \midrule
    MoCov2~\cite{chen2020improved} & 800 & 71.1  \\
    SimCLR~\cite{chen2020simple} & 1000 & 69.3 \\ 
    BYOL~\cite{jean-bastien2020bootstrap} & 1000 & 74.3  \\
    SimSiam~\cite{chen2021exploring} & 800 & 71.3 \\
    Barlow Twins~\cite{zbontar2021barlow} & 1000 & 73.2 \\
    VICReg~\cite{bardes2021vicreg} & 1000 & 73.2 \\
    DINO (+multi-crop)~\cite{caron2021emerging} & 800 & 75.3 \\
    TWIST (+multi-crop)~\cite{wang2021self} & 800 & 75.5 \\
    \midrule
    \name+CutMix & 800 & 74.9 \\
    \name+CutMix+multi-crop & 800 & 75.5 \\
    \bottomrule
    \end{tabular}}
    \caption{Linear classification on ImageNet~\cite{deng2009imagenet}.}
    \label{tab:sota}
    \vspace{-1.3em}
\end{table}
\section{Conclusion}
\label{sec:conclusion}
In this paper, we present a unified framework for three typical self-supervised learning methods from the perspective of gradient analysis. While previous works appear to be distinct in their loss functions, we demonstrate that they share a similar gradient form. Such form consists of the positive gradient, the negative gradient and the balance factor, which suggests that these methods work in a similar mechanism. We further compare their performances under a fair experiment setting. It's shown that they can deliver similar performances, and momentum encoder is the key factor to boost performance. Finally, we propose \name, a simple but effective gradient form for self-supervised learning. Extensive experiments have shown its effectiveness in linear evaluation and downstream tasks.

\vspace{0.3em}
\noindent\textbf{Limitations.} This work only adopts linear evaluation for performance comparison, while different methods may have a different impact on downstream tasks, \eg, object detection and semantic segmentation. We leave the transfer learning performance comparison for future work.

\vspace{0.3em}
\noindent\textbf{Potential Negative Societal Impact.} This work may inherit the negative impacts of self-supervised learning. Because a large-scale training is usually required, it may consume lots of electricity and cause environmental pollution. This method also learns representations from training dataset and may contain data biases. Future work can seek for a more efficient and unbiased training method.
\paragraph{Acknowledgments} The work is supported by the National Key R\&D Program of China (2020AAA0105200), Beijing Academy of Artificial Intelligence.

{\small
\bibliographystyle{ieee_fullname}
\bibliography{egbib.bib}
}

\clearpage
\appendix
\section{Gradient Analysis}
\label{appendix:grad_analysis}
\subsection{Contrastive Learning Methods}

\noindent\textbf{Derivation of the gradient for MoCo~\cite{he2020momentum}}. For simplicity, we denote $l(u_1^o)$ as the InfoNCE loss for the sample $u_1^o$:
\begin{equation}
    \label{eq: infonce_moco}
	l(u_1^o) = -\log\frac{\exp{(\mathrm{cos}(u_1^o, u_2^m)/\tau)}}{\sum_{v^m\in\mathcal{V}_\mathrm{bank}}\exp{(\mathrm{cos}(u_1^o, v^m)/\tau)}}.
\end{equation}

Let $l(u_1^o) = -\mathrm{log}s_{u_2}$, $s_{u_2}= \frac{\mathrm{exp}(c_{u_2^m})}{\sum_{v^m\in\mathcal{V}_\mathrm{bank}} \mathrm{exp}(c_{v^m})}$, $c_{v} = \mathrm{cos}(u_1^o, v) / \tau$. According to the chain rule, we have
\begin{equation}
\begin{split}
    \frac{\partial l(u_1^o)}{\partial u_1^o} 
    ={}& \frac{\partial l(u_1^o)}{\partial s_{u_2}} \cdot \frac{\partial s_{u_2}}{\partial c_{u_2^m}} \cdot \frac{\partial c_{u_2^m}}{\partial u_1^o} \\
    & + \sum_{v^m\in\mathcal{V}_\mathrm{bank}\setminus u_2^m}\frac{\partial l(u_1^o)}{\partial s_{u_2}} \cdot \frac{\partial s_{u_2}}{\partial c_{v^m}} \cdot \frac{\partial c_{v^m}}{\partial u_1^o} \\
    =& -\frac{1}{s_{u_2}} \cdot s_{u_2}(1-s_{u_2}) \cdot \frac{u_2^m}{\tau} \\
    & - \sum_{v^m\in\mathcal{V}_\mathrm{bank}\setminus u_2^m}\frac{1}{s_{u_2}} \cdot s_{u_2}s_v \cdot \frac{v^m}{\tau} \\
    =& -\frac{u_2^m}{\tau} + \sum_{v^m\in\mathcal{V}_\mathrm{bank}}s_v \frac{v^m}{\tau},
\end{split}    
\end{equation}
where $s_v=\frac{\exp{(\mathrm{cos}(u_1^o, v^m)/\tau)}}{\sum_{y^m\in \mathcal{V}_{\mathrm{bank}}}\exp{(\mathrm{cos}(u_1^o, y^m)/\tau)}}$.

Denote $L$ as the averaged $l(\cdot)$ over a batch of N samples, its gradient w.r.t $u_1^o$ is 
\begin{equation}
    \frac{\partial L}{\partial u_1^o} = \frac{1}{N} \frac{\partial l(u_1^o)}{\partial u_1^o} =  \frac{1}{\tau N}\bigg(-u_2^{m} + \sum_{v^{m}\in\mathcal{V}_{\mathrm{bank}}}s_vv^m\bigg).
\end{equation}

\vspace{-0.6em}
\begin{algorithm}[h]
\caption{Pseudocode of MoCo in PyTorch style.}
\label{alg:code}
\definecolor{codeblue}{rgb}{0.25,0.5,0.5}
\lstset{
  backgroundcolor=\color{white},
  basicstyle=\fontsize{7.2pt}{7.2pt}\ttfamily\selectfont\linespread{0.8},
  columns=fullflexible,
  breaklines=true,
  captionpos=b,
  commentstyle=\fontsize{7.2pt}{7.2pt}\color{codeblue},
  keywordstyle=\fontsize{7.2pt}{7.2pt},
}
\vspace{-0.8em}
\begin{lstlisting}[language=python]
# u1, u2: normalized representations for two augmented views of shape [N, C]
# V_bank: the memory bank of shape [K, C]
# tau: the temperature coefficient 

# positive term
loss_pos = -(u1*u2.detach()).sum(-1) # [N, 1] 
# negative term
weight = softmax(u1@V_bank.T/tau, dim=-1) # [N, K]
loss_neg = (weight@V_bank)*v1.sum(-1) # [N, 1]
# MoCo
loss = 1/tau * (loss_pos + loss_neg).mean()

\end{lstlisting}
\vspace{-0.8em}
\end{algorithm}

\noindent\textbf{Derivation of the gradient for SimCLR~\cite{chen2020simple}}. For SimCLR, the InfoNCE loss $l(u_1^o)$ should be modified as 
\begin{equation}
	l(u_1^o) = -\log\frac{\exp{(\mathrm{cos}(u_1^o, u_2^s)/\tau)}}{\sum_{v^s\in\mathcal{V}_\mathrm{batch}\setminus u_1^o}\exp{(\mathrm{cos}(u_1^o, v^s)/\tau)}}.
\end{equation}

Note that because the target branch is not detached from back-propagation, $u_1^o$ can receive gradients from $l(u_2^s)$ and $l(v^s)$. Accordingly, the gradient can be derived as 
\allowdisplaybreaks
{\small
\begin{align}
\label{eq:simclr_full_a}
    \frac{\partial L}{\partial u_1^o} ={}& \frac{1}{N}\bigg( \frac{\partial l(u_1^o)}{\partial u_1^o} + \frac{\partial l(u_2^s)}{\partial u_1^o} + \sum_{v^{s}\in\mathcal{V}_{\mathrm{batch}}\setminus\{u_1^o, u_2^s\}}\frac{\partial l(v^s)}{\partial u_1^o}\bigg) \nonumber\\
	={}&\frac{1}{\tau N}\bigg(-u_2^{s} + \sum_{v^{s}\in\mathcal{V}_{\mathrm{batch}}\setminus u_1^o}s_vv^s\bigg) \nonumber\\
	&+\frac{1}{\tau N}\bigg(-u_2^{s} + t_{u_2}u_2^s\bigg)
	+\frac{1}{\tau N}\sum_{v^{s}\in\mathcal{V}_{\mathrm{batch}}\setminus \{u_1^o,u_2^s\}}t_vv^s \nonumber\\
	={}&\frac{1}{\tau N}\bigg(-u_2^{s} + \sum_{v^{s}\in\mathcal{V}_{\mathrm{batch}}\setminus u_1^o}s_vv^s\bigg) \nonumber\\
	&+\frac{1}{\tau N}\bigg(-u_2^{s} + \sum_{v^{s}\in\mathcal{V}_{\mathrm{batch}}\setminus u_1^o}t_vv^s\bigg),
\end{align}}
where $t_v=\frac{\exp{(\mathrm{cos}(v^s, u_1^o)/\tau)}}{\sum_{y^s\in \mathcal{V}_{\mathrm{batch}}\setminus v^s}\exp{(\mathrm{cos}(v^s, y^s)/\tau)}}$. If we stop the gradient from $l(u_2^s)$ and $l(v^s)$, Eq.(\ref{eq:simclr_full_a}) will reduce to
\begin{align}
\label{eq:simclr_sim_a}
    \frac{\partial L}{\partial u_1^o} 
	\approx{}&\frac{1}{\tau N}\bigg(-u_2^{s} + \sum_{v^{s}\in\mathcal{V}_{\mathrm{batch}}\setminus u_1^o}s_vv^s\bigg),
\end{align}
which shares a similar structure with that of MoCo. We demonstrate empirically that this simplification dose no harm to the performance as shown in Table~\ref{tab:grad_sim}.

\vspace{-0.6em}
\begin{algorithm}[h]
\caption{Pseudocode of Simplified SimCLR in PyTorch style.}
\label{alg:code}
\definecolor{codeblue}{rgb}{0.25,0.5,0.5}
\lstset{
  backgroundcolor=\color{white},
  basicstyle=\fontsize{7.2pt}{7.2pt}\ttfamily\selectfont\linespread{0.8},
  columns=fullflexible,
  breaklines=true,
  captionpos=b,
  commentstyle=\fontsize{7.2pt}{7.2pt}\color{codeblue},
  keywordstyle=\fontsize{7.2pt}{7.2pt},
}
\vspace{-0.8em}
\begin{lstlisting}[language=python]
# u1, u2: normalized representations for two augmented views of shape [N, C]
# tau: the temperature coefficient 

# positive term
loss_pos = -(u1*u2.detach()).sum(-1) # [N, 1] 
# negative term
weight = softmax(u1@u2.T/tau, dim=-1) # [N, N]
loss_neg = (weight@u2).detach()*u1.sum(-1) # [N, 1]
# simplified SimCLR
loss = 1/tau * (loss_pos + loss_neg).mean()

\end{lstlisting}
\vspace{-0.8em}
\end{algorithm}

\subsection{Asymmtric Network Methods}
\noindent\textbf{Derivation of the gradient for DirectPred~\cite{tian2021understanding}}. DirectPred takes the negative cosine similarity loss between target sample and projected online sample:
\begin{equation}
	l(u_1^o) = -\mathrm{cos}(\frac{W_hu_1^o}{||W_h u_1^o||_2}, u_2^t),
\end{equation}
\begin{equation}
    W_h = U\Lambda_h U^T, \ \ \Lambda_h = \Lambda_{F}^{1/2}+\epsilon \lambda_{max}I,
\end{equation}
where $U$ and $\Lambda_F$ are the eigenvectors and eigenvalues of $F= \sum_{v^o \in \mathcal{V}_{\infty}} \rho_vv^o {v^o}^T$, respectively. $\epsilon$ is a hyperparameter to boost small eigenvalues.

\begin{figure}[t]
\begin{subfigure}{.23\textwidth}
    \centering
    \includegraphics[width=1.0\textwidth]{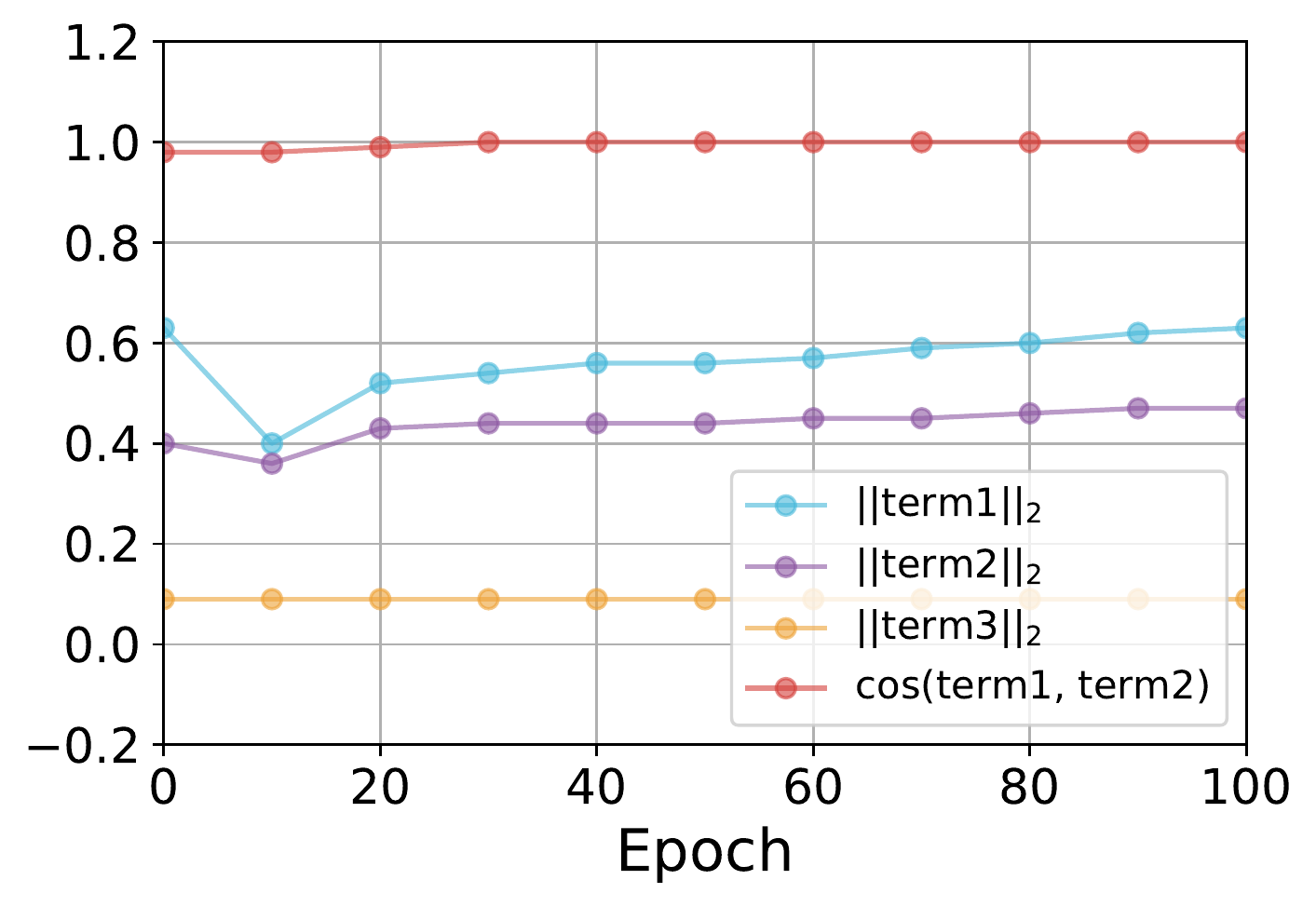}
    \subcaption{Dynamics of three terms in Eq.~(\ref{eq:asym_full_b}) (Appendix)}
\end{subfigure}
\begin{subfigure}{.23\textwidth}
    \centering
    \includegraphics[width=1.0\textwidth]{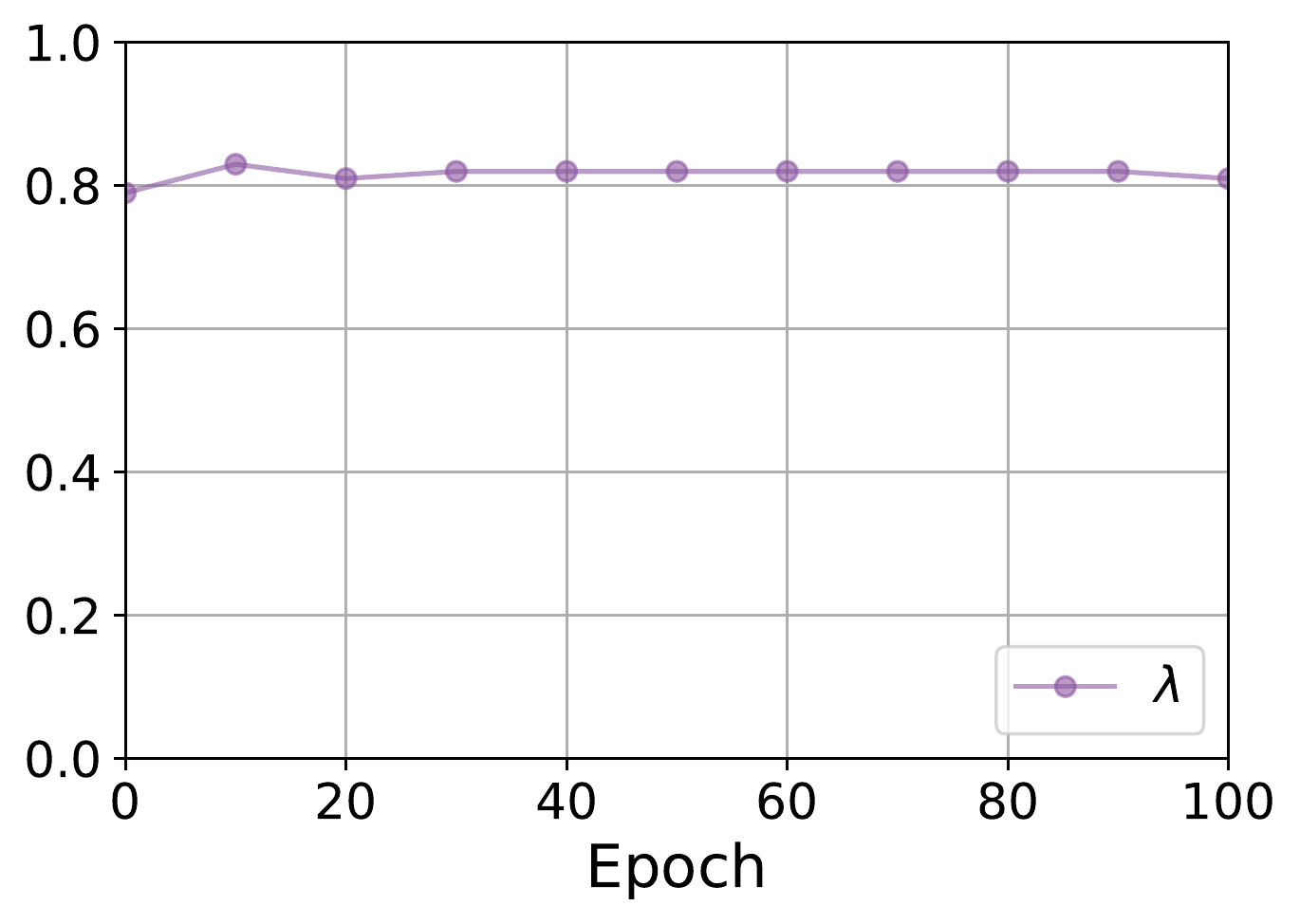}
    \subcaption{Dynamics of ${\tilde{\lambda}}$ in Eq.~(\ref{eq:asym_full_b}) (Appendix)}
\end{subfigure}
\begin{subfigure}{.23\textwidth}
    \centering
    \includegraphics[width=1.0\textwidth]{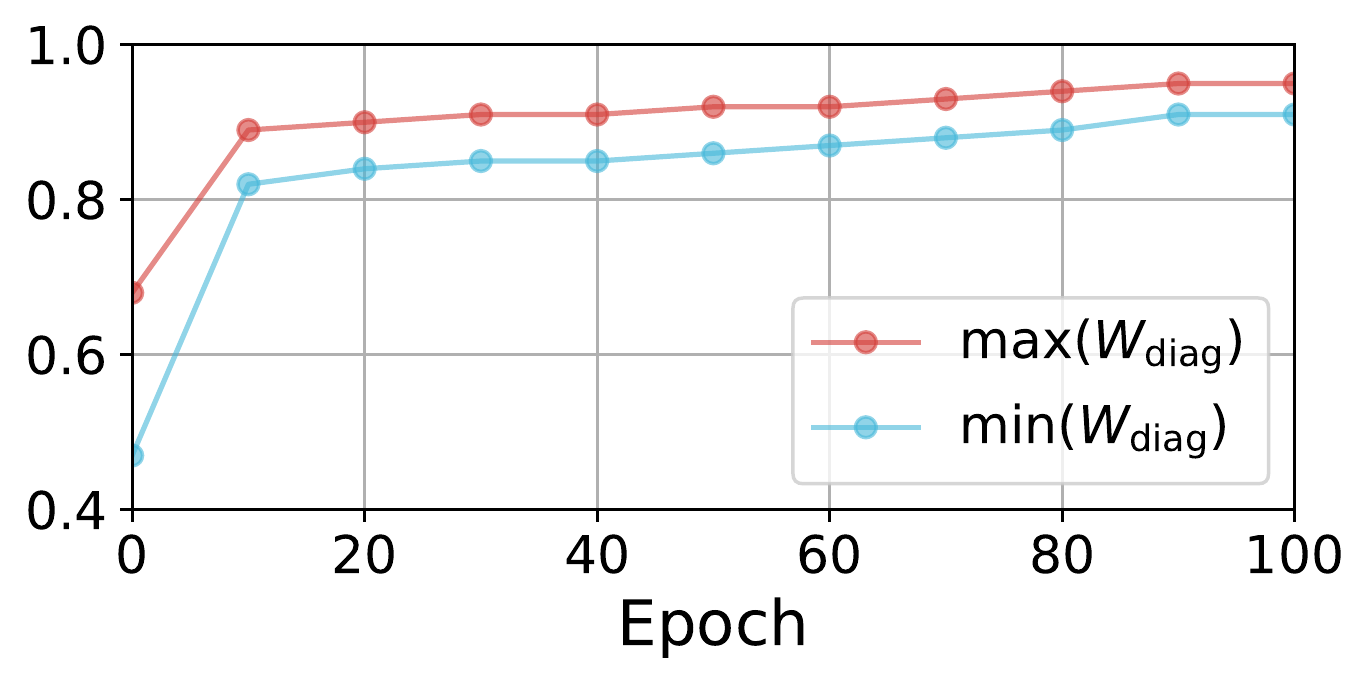}
    \subcaption{Dynamics of the max and min value of $W_{\mathrm{diag}}$ in Eq.~(\ref{eq:bt_grad})}
\end{subfigure}
\hspace{0.1em}
\begin{subfigure}{.23\textwidth}
    \centering
    \includegraphics[width=1.0\textwidth]{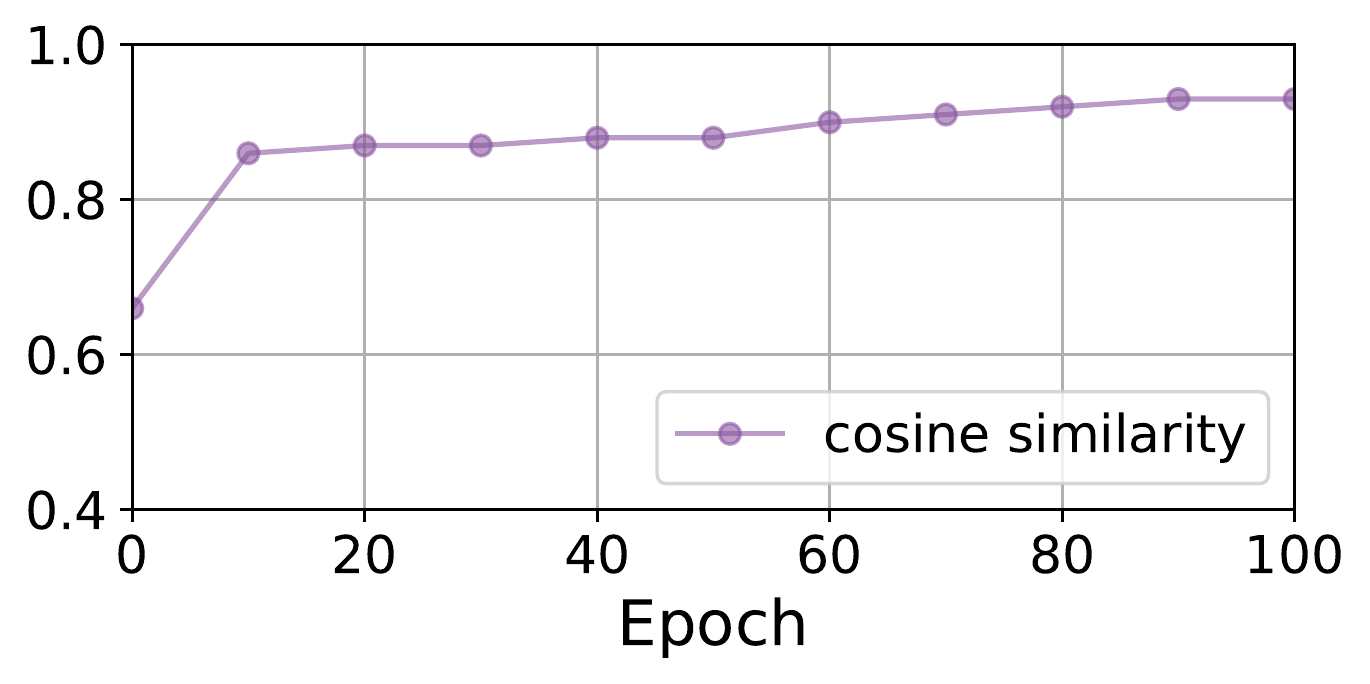}
    \subcaption{Dynamics of similarity between the neglected and negative terms in Eq.~(\ref{eq:vic_grad})}
\end{subfigure}
    \caption{Justifications for simplifications.}
    \label{fig:justification}
\end{figure}

Denote $y_1 = W_hu_1^o$, $y_1^n = \frac{y_1}{||y_1||_2}$, the gradient of $L$ can be derived as: 
{\small
\begin{align}
    \label{eq:asy_full_a}
    \frac{\partial L}{\partial u_1^o} ={}& \frac{1}{N}\bigg( 
    \frac{\partial y_1}{\partial u_1^o}\cdot
    \frac{\partial y_1^n}{\partial y_1}\cdot
    \frac{\partial l}{\partial y_1^n}
    \bigg) \nonumber\\
	={}&  \frac{1}{N}\bigg(- W_h^T\cdot
    \frac{1}{||y_1||_2} (I-\frac{y_1 y_1^T}{y_1^T y_1})\cdot u_2^t
	\bigg) \nonumber\\
	={}& \frac{1}{||W_h u_1^o||_2N}\bigg(-W_h^T (I-\frac{W_h u_1^o {u_1^o}^T W_h^T}{{u_1^o}^T W_h^T W_h u_1^o}) u_2^t \bigg) \nonumber\\ 
	={}& \frac{1}{||W_h u_1^o||_2N}\bigg(-W_h^T u_2^t + \frac{{u_1^o}^T W_h^Tu_2^t}{{u_1^o}^T W_h^T W_h u_1^o} W_h^T W_h u_1^o \bigg).
\end{align}}
Note that 
\begin{align}
\label{eq:asym_wh}
    W_h^T W_h ={}& U\Lambda_h^TU^TU\Lambda_hU^T \nonumber \\
    ={}& U(\Lambda_F + 2\epsilon\lambda_{max}\Lambda_F^{1/2}+\epsilon^2\lambda_{max}^2I)U^T \nonumber \\
    ={}& F+2\epsilon\lambda_{max}F^{1/2}+\epsilon^2\lambda_{max}^2I.
\end{align}
Substituting Eq.(\ref{eq:asym_wh}) into Eq.(\ref{eq:asy_full_a}) leads to 
{\small
\begin{align}
\label{eq:asym_full_b}
    \frac{\partial L}{\partial u_1^o}
    ={}& \frac{1}{||W_h u_1^o||_2N}\bigg(-W_h^T u_2^t \nonumber\\
    & + \tilde{\lambda} (Fu_1^o+2\epsilon\lambda_{max}F^{1/2}u_1^o+\epsilon^2\lambda_{max}^2u_1^o)  \bigg),
\end{align}}
where $\tilde{\lambda}=\frac{{u_1^o}^T W_h^Tu_2^t}{{u_1^o}^T(F+2\epsilon\lambda_{max}F^{1/2}+\epsilon^2\lambda_{max}^2I) u_1^o}$. For those three terms that are scaled by $\tilde{\lambda}$, we plot the value of their magnitude and the similarity of the first two terms in Figure ~\ref{fig:justification}(a). It's shown that the first two terms have highly similar direction so they are expected to have similar effect on the training. We have also verified that removing the $F^{1/2}$ term will not cause performance drop (see Table~\ref{tab:grad_sim}). Thus, the gradient can be simplified into
{\small
\begin{align}
\label{eq:asym_sim}
    \frac{\partial L}{\partial u_1^o}
    \approx{}& \frac{1}{||W_h u_1^o||_2N}\bigg(-W_h^T u_2^t + \lambda (Fu_1^o+\epsilon^2\lambda_{max}^2u_1^o)  \bigg),
\end{align}}
where $\lambda=\frac{{u_1^o}^T W_h^Tu_2^t}{{u_1^o}^T(F+\epsilon^2\lambda_{max}^2I) u_1^o}$.

When $u_1^o$ is $\ell_2$ normalized, we can further neglect the $\epsilon^2\lambda_{max}^2u_1^o$ term, because the gradient propagated to unnormalized  $u_1^o$ is $0$. Hence, we simplify the gradient as 
\begin{align}
    \frac{\partial L}{\partial u_1^o} 
    \approx{}& \frac{1}{||W_h u_1^o||_2N}\bigg(-W_h^T u_2^t 
    + \lambda Fu_1^o  \bigg) \nonumber\\
    ={}& \frac{1}{||W_h u_1^o||_2N}\bigg(-W_h^T u_2^t+\lambda\sum_{v^o \in \mathcal{V}_{\infty}} (\rho_v{u_1^o}^Tv^o)v^o\bigg).
\end{align}
Note that $\lambda$ is a dynamic balance factor, but we find that its value tends to be quite stable (see Figure~\ref{fig:justification}(b)), so it can also be substiuted by a constant scalar.

\vspace{-0.6em}
\begin{algorithm}[ht]
\caption{Pseudocode of Asymetric Networks.}
\label{alg:code}
\definecolor{codeblue}{rgb}{0.25,0.5,0.5}
\lstset{
  backgroundcolor=\color{white},
  basicstyle=\fontsize{7.2pt}{7.2pt}\ttfamily\selectfont\linespread{0.8},
  columns=fullflexible,
  breaklines=true,
  captionpos=b,
  commentstyle=\fontsize{7.2pt}{7.2pt}\color{codeblue},
  keywordstyle=\fontsize{7.2pt}{7.2pt},
}
\vspace{-0.8em}
\begin{lstlisting}[language=python]
# u1, u2: normalized representations for two augmented views of shape [N, C]
# F: the moving average of correlation matrix
# rho: the moving average coefficient
# eps: hyperparameter to boost small eigenvalues

# accumulate F
tmp_F = ((u1.T@u1 + u2.T@u2) / (2*N)).detach()
F = rho*F + (1-rho)*tmp_F # update moving average

# calculate Wh
U, lambda_F, V = torch.svd(F)
lambda_h = torch.sqrt(lambda_F) + eps*lambda_F.max()
Wh = U@(torch.diag(lambda_h)@V) # [C, C]

# positive term
loss_pos = -(u1*(u2@Wh).detach()).sum(-1) # [N, 1]
# negative term
loss_neg = (u1*(u1@F).detach()).sum(-1) # [N, 1]

weight = 1/torch.linalg.norm(u1@Wh, dim=-1)
loss = (weight*(loss_pos + lambda * loss_neg)).mean()

\end{lstlisting}
\vspace{-0.8em}
\end{algorithm}

\begin{table}[h]
  \centering
  \begin{tabular}{@{}ccccc@{}}
    \toprule
    Method & \multicolumn{2}{c}{SimCLR~\cite{chen2020simple}} & \multicolumn{2}{c}{DirectPred~\cite{tian2021understanding}} \\
    \midrule
    Gradient & Eq.(\ref{eq:simclr_full_a}) & Eq.(\ref{eq:simclr_sim_a}) & Eq.(\ref{eq:asym_full_b}) & Eq.(\ref{eq:asym_sim}) \\ 
    Linear Eval & 67.5 & 67.6 & 70.2 & 70.2 \\
    \bottomrule
  \end{tabular}
  \caption{Simplification for the gradient of SimCLR and DirectPred. We use the 100-epoch pre-training and lineal evaluation protocol described in Appendix~\ref{appendix:implementation}.}
  \label{tab:grad_sim}
\end{table}

\subsection{Feature Decorrelation Methods}
\noindent\textbf{Derivation of the gradient for Barlow Twins~\cite{zbontar2021barlow}}. Barlow Twins forces the cross-correlation matrix to be close to the identity matrix via the following loss function:
\begin{equation}
    \label{eq: loss_bt_a}
    L = \sum_{i=1}^C(W_{ii}-1)^2+\lambda\sum_{i=1}^C\sum_{j\not=i}W_{ij}^2,
\end{equation}
where $W=\frac{1}{N}\sum_{v_1^o, v_2^s\in\mathcal{V}_{\mathrm{batch}}}v_1^ov_2^{sT}$ is the cross-correlation matrix.

Denote $L_1=\sum_{i=1}^C(W_{ii}-1)^2$, $L_2=\lambda\sum_{i=1}^C\sum_{j\not=i}W_{ij}^2$. We use the operator $(\cdot)_k$ to represent the $k$-th element of a vector. For $L_1$, We have:
\begin{align}
    \label{eq: grad_bt_l1_a}
        \frac{\partial L_1}{\partial (u_1^o)_k} = \frac{\partial L_1}{\partial W_{kk}} \cdot \frac{\partial W_{kk}}{\partial (u_1^o)_k} = 2(W_{kk}-1) \cdot \frac{(u_2^s)_k}{N}.
\end{align}

For $L_2$, we have:
{\small
\begin{align}
    \label{eq: grad_bt_l2_a}
        \frac{\partial L_2}{\partial (u_1^o)_k} &{}= \lambda\sum_{j\not=k}^C 2 \frac{\partial L_2}{\partial W_{kj}} \cdot \frac{\partial W_{kj}}{\partial (u_1^o)_k} =\lambda\sum_{j\not=k}^C 2 W_{kj} \cdot \frac{(u_2^s)_j}{N} \nonumber\\
        =& \frac{2\lambda}{N} \bigg(- W_{kk}(u_2^s)_k+\sum_{j=1}^C W_{kj} (u_2^s)_j\bigg) \nonumber\\
        =& \frac{2\lambda}{N} \bigg(- W_{kk}(u_2^s)_k+\sum_{j=1}^C \frac{1}{N} \sum_{v_1^o, v_2^s\in\mathcal{V}_{\mathrm{batch}}}(v_1^o)_k (v_2^{s})_j (u_2^s)_j\bigg) \nonumber \\
        =& \frac{2\lambda}{N} \bigg(- W_{kk}(u_2^s)_k+\sum_{v_1^o, v_2^s\in\mathcal{V}_{\mathrm{batch}}} \frac{1}{N} (v_1^o)_k \sum_{j=1}^C (v_2^{s})_j (u_2^s)_j\bigg) \nonumber\\
        =& \frac{2\lambda}{N} \bigg(- W_{kk}(u_2^s)_k+\sum_{v_1^o, v_2^s\in\mathcal{V}_{\mathrm{batch}}} \frac{u_2^{sT}v_2^s}{N} (v_1^o)_k\bigg).
    \raisetag{1.8\baselineskip}
\end{align}}
Combining Eq.(\ref{eq: grad_bt_l1_a}) and Eq.(\ref{eq: grad_bt_l2_a}) together, we get: 
\begin{align}
       \frac{\partial L}{\partial u_1^o} =  \frac{2}{N} \bigg(-Au_2^s+\lambda\sum_{v_1^o, v_2^s\in\mathcal{V}_{\mathrm{batch}}}{\frac{u_2^{sT}v_2^s}{N}v_1^o}\bigg),
\end{align}
where $A=I-(1-\lambda)W_{\mathrm{diag}}$. Here $(W_{\mathrm{diag}})_{ij} = \delta_{ij} W_{ij}$ is the diagonal matrix of $W$, where $\delta_{ij}$ is the Kronecker delta.

\vspace{-0.6em}
\begin{algorithm}[h]
\caption{Pseudocode of Barlow Twins in PyTorch style.}
\label{alg:code}
\definecolor{codeblue}{rgb}{0.25,0.5,0.5}
\lstset{
  backgroundcolor=\color{white},
  basicstyle=\fontsize{7.2pt}{7.2pt}\ttfamily\selectfont\linespread{0.8},
  columns=fullflexible,
  breaklines=true,
  captionpos=b,
  commentstyle=\fontsize{7.2pt}{7.2pt}\color{codeblue},
  keywordstyle=\fontsize{7.2pt}{7.2pt},
}
\vspace{-0.8em}
\begin{lstlisting}[language=python]
# u1, u2: representations for two augmented views of shape [N, C]
# lambda: the moving average coefficient

# correlation matrix 
W_cor = u1.T@u2 / N # [C, C]
# positive term
pos = (1 - (1-lambda)*torch.diag(W_cor)) * u2
loss_pos = -(u1 * pos.detach()).sum(-1) # [N, 1] 
# negative term
weight = u2@u2.T / N 
loss_neg = ((weight@u1).detach() * u1).sum(-1)
# Barlow Twins
loss = 2 * (loss_pos + lambda * loss_neg).mean()

\end{lstlisting}
\vspace{-0.8em}
\end{algorithm}

\vspace{0.5em}
\noindent\textbf{Derivation of the gradient for VICReg~\cite{bardes2021vicreg}}. The loss function of VICReg consists of three componets: 
\begin{equation}
    L_1 = \frac{1}{N}\sum_{v_1^o, v_2^s\in\mathcal{V}_{\mathrm{batch}}}||v_1^o-v_2^s||_2^2,
\end{equation}
\begin{equation}
    L_2 = \frac{\lambda_1}{C}\sum_{i=1}^C\sum_{j\ne i}^CW_{ij}'^2,
\end{equation}
\begin{equation}
    L_3 = \frac{\lambda_2}{C}\sum_{i=1}^C\max(0, \gamma - \mathrm{std}(v_1^o)_i),
\end{equation}
where $W'=\frac{1}{N-1}\sum_{v_1^o\in\mathcal{V}_{\mathrm{batch}}}(v_1^o-\bar{v}_1^o)(v_1^o-\bar{v}_1^o)^{T}$.

For the invariance term $L_1$, we have: 
\begin{equation}
        \frac{\partial L_1}{\partial (u_1^o)_k} = \frac{2}{N}(u_1^o-u_2^s)_k.
\end{equation}

For the covariance term $L_2$, we have: 
{\small
\begin{align}
        \frac{\partial L_2}{\partial (u_1^o)_k} &={} \frac{2\lambda_1}{C}\sum_{j\not=k}^C \frac{\partial L_2}{\partial W_{kj}'}\cdot \frac{\partial W_{kj}'}{\partial (u_1^o)_k} \nonumber\\
        ={}& \frac{4\lambda_1}{C}\sum_{j\not=k}^C W_{kj}'\frac{(u_1^o-\bar{v}_1^o)_j}{N-1} \nonumber\\
        ={}& \frac{4\lambda_1}{C(N-1)}\bigg(-W_{kk}'(u_1^o-\bar{v}_1^o)_k+\sum_{j=1}^C W_{kj}'(u_1^o-\bar{v}_1^o)_j\bigg) \nonumber\\
        ={}& \frac{4\lambda_1}{C(N-1)}\bigg(-W_{kk}'(u_1^o-\bar{v}_1^o)_k \nonumber\\
        &+\sum_{j=1}^C\sum_{v_1^o\in\mathcal{V}_{\mathrm{batch}}}\frac{(u_1^o-\bar{v}_1^o)^T(v_1^o-\bar{v}_1^o)}{N-1} (v_1^o-\bar{v}_1^o)_j\bigg) \nonumber\\
        =& \frac{4\lambda_1N}{C(N-1)^2}\bigg(-\frac{N-1}{N}W_{kk}'(\tilde{u}_1^o)_k +\sum_{v_1^o\in\mathcal{V}_{\mathrm{batch}}}\frac{\tilde{u}_1^{oT}\tilde{v}_1^o}{N} (\tilde{v}_1^o)_j\bigg) \nonumber\\
        =& \frac{2\lambda}{N}\bigg(-\frac{N-1}{N}W_{kk}'(\tilde{u}_1^o)_k +\sum_{v_1^o\in\mathcal{V}_{\mathrm{batch}}}\frac{\tilde{u}_1^{oT}\tilde{v}_1^o}{N} (\tilde{v}_1^o)_j\bigg),
\end{align}}
where $\lambda=\frac{2\lambda_1N^2}{C(N-1)^2}$ and $\tilde{v}=v-\bar{v}$ is the de-centered sample. 

For the variance term $L_3$, we have: 
\begin{align}
        \frac{\partial L_3}{\partial (u_1^o)_k} ={}& \frac{\lambda_2}{C} \frac{\partial \max(0, \gamma - \mathrm{std}(v_1^o)_k)}{\partial \mathrm{std}(v_1^o)_k}\cdot \frac{\partial \mathrm{std}(v_1^o)_k}{\partial (u_1^o)_k} \nonumber\\
        ={}& -\frac{\lambda_2}{C(N-1)}\mathbbm{1}(\gamma-\mathrm{std}(v_1^o)_k>0) \frac{(\tilde{u}_1^o)_k}{\mathrm{std}(v_1^o)_k}.
\end{align}

For final loss function $L = L_1+L_2+L_3$, its gradient w.r.t $u_1^o$ can be represented as: 
\begin{equation}
\small
    \begin{split}
        \frac{\partial L}{\partial u_1^o} ={}& \frac{2}{N}(u_1^o-u_2^s)  -\frac{2\lambda}{N}\bigg(\frac{N-1}{N}W_{\mathrm{diag}}'\tilde{u}_1^o -\sum_{v_1^o\in\mathcal{V}_{\mathrm{batch}}}\frac{\tilde{u}_1^{oT}\tilde{v}_1^o}{N} \tilde{v}_1^o\bigg) \\
        &-\frac{\lambda_2}{C(N-1)}\mathrm{diag}(\mathbbm{1}(\gamma-\mathrm{std}(v_1^o)>0)\oslash\mathrm{std}(v_1^o)) \tilde{u}_1^o \\
        ={}& \frac{2}{N}\bigg(-u_2^s+\lambda\sum_{v_1^o\in\mathcal{V}_{\mathrm{batch}}}\frac{\tilde{u}_1^{oT}\tilde{v}_1^o}{N} \tilde{v}_1^o\bigg)
        +\frac{2\lambda}{N}\bigg(\frac{1}{\lambda}u_1^o- B\tilde{u}_1^o\bigg), \\
    \end{split}
\end{equation}

\vspace{-0.6em}
\begin{algorithm}[t]
\caption{Pseudocode of Simplified VICReg in PyTorch style.}
\label{alg:code}
\definecolor{codeblue}{rgb}{0.25,0.5,0.5}
\lstset{
  backgroundcolor=\color{white},
  basicstyle=\fontsize{7.2pt}{7.2pt}\ttfamily\selectfont\linespread{0.8},
  columns=fullflexible,
  breaklines=true,
  captionpos=b,
  commentstyle=\fontsize{7.2pt}{7.2pt}\color{codeblue},
  keywordstyle=\fontsize{7.2pt}{7.2pt},
}
\vspace{-0.8em}
\begin{lstlisting}[language=python]
# u1, u2: representations for two augmented views of shape [N, C]
# lambda: the moving average coefficient

# positive term
loss_pos = -(u1 * u2.detach()).sum(-1) # [N, 1] 
# negative term
weight = u1@u1.T / N 
loss_neg = ((weight@u1).detach() * u1).sum(-1)
# simplified VICReg
loss = 2 * (loss_pos + lambda * loss_neg).mean()

\end{lstlisting}
\vspace{-0.8em}
\end{algorithm}

where $B=\frac{N}{\lambda C(N-1)}(2\lambda_1W'_{\mathrm{diag}} + \frac{\lambda_2}{2}\mathrm{diag}(\mathbbm{1}(\gamma-\mathrm{std}(v_1^o)>0)\oslash\mathrm{std}(v_1^o)))$. Here $W_{\mathrm{diag}}'$ is the diagonal matrix of $W'$, $\mathrm{diag}(x)$ is a matrix with diagonal filled with the vector $x$, $\mathbbm{1}(\cdot)$ is the indicator function, and $\oslash$ denotes element-wise division.

\subsection{Pseudocode of UniGrad}

\vspace{-0.6em}
\begin{algorithm}[h]
\caption{Pseudocode of UniGrad in PyTorch style.}
\label{alg:code}
\definecolor{codeblue}{rgb}{0.25,0.5,0.5}
\lstset{
  backgroundcolor=\color{white},
  basicstyle=\fontsize{7.2pt}{7.2pt}\ttfamily\selectfont\linespread{0.8},
  columns=fullflexible,
  breaklines=true,
  captionpos=b,
  commentstyle=\fontsize{7.2pt}{7.2pt}\color{codeblue},
  keywordstyle=\fontsize{7.2pt}{7.2pt},
}
\vspace{-0.8em}
\begin{lstlisting}[language=python]
# u1, u2: normalized representations for two augmented views of shape [N, C]
# F: the moving average of correlation matrix
# rho: the moving average coefficient

# positive term
loss_pos = -(u1*u2.detach()).sum(-1) # [N, 1] 
# negative term
tmp_F = ((u1.T@u1 + u2.T@u2) / (2*N)).detach()
F = rho*F + (1-rho)*tmp_F # update moving average
loss_neg = (u1@F)*u1.sum(-1) # [N, 1]
# UniGrad
loss = (loss_pos + lambda * loss_neg).mean()

\end{lstlisting}
\vspace{-0.8em}
\end{algorithm}

\section{Implementation Details}
\label{appendix:implementation}
We provide the experimental settings used in this paper. For 100 epochs pre-training and linear evaluation, we mainly follow \cite{chen2021exploring}; For 800 epochs pre-training, large batch size is adopted for faster training and hence we mainly follow \cite{jean-bastien2020bootstrap}.

\vspace{0.5em}
\noindent\textbf{Pre-training setting for 100 epochs.} SGD is used as the optimizer. The weight decay is $1.0\times10^{-4}$ and the momentum is $0.9$. The learning rate is set according to linear scaling rule~\cite{goyal2017accurate} as $base\_lr\times batch\_size/256$, with $base\_lr=0.05$. The learning rate has a cosine decay schedule for 100 epochs with 5 epochs linear warmup. The batch size is set to $1024$. We use ResNet50~\cite{he2016deep} as the backbone. The projection MLP has three layers, with the hidden and output dimension set to $2048$. BN and ReLU are applied after the first two layers. If a momentum encoder is used, we follow  BYOL~\cite{jean-bastien2020bootstrap} to increase the exponential moving average parameter from $0.996$ to $1$ with a cosine scheduler. We list these hyper-parameters in Table~\ref{tab:hyper}.

\begin{table}[h]
    \centering
    \begin{tabular}{@{}ll@{}}
    \toprule
        Hyper-parameter & Value \\
    \midrule
        opitmizer & SGD \\
        weight decay & $1.0\times10^{-4}$ \\
        base lr & $0.05$ \\
        lr schedule & cosine \\
        warmup & 5 epochs \\
        batch size & 1024 \\
        projector & 3-layers MLP \\
        init momentum & $0.996$ \\
        final momentum & $1.0$ \\
        momentum schedule & cosine \\
    \bottomrule
    \end{tabular}
    \caption{Hyper-parameters for 100 epochs pre-training. We use the same hyper-parameters for different loss functions.}
    \label{tab:hyper}
\end{table}

\vspace{0.5em}
\noindent\textbf{Pre-training setting for 800 epochs.} LARS~\cite{you2017scaling} optimizer is used for 800 epochs pre-training with a batch size of $4096$. The weight decay is $1.0\times10^{-6}$ and the momentum is $0.9$. The learning rate is set with $base\_lr=0.3$ for the weights and $base\_lr=0.05$ for the biases and batch normalization parameters. Cosine decay schedule is used after a linear warm-up of 10 epochs. We exclude the biases and batch normalization parameters from the LARS adaptation and weight decay. For the projector, We use a three-layer MLP with hidden and output dimension set to $8192$. Other configurations keep the same as the pre-training setting for 100 epochs.

\vspace{0.5em}
\noindent\textbf{Linear evaluation.} We follow the common practice to adopt linear evaluation as the performance metric. Such practice trains a supervised linear classifier on top of the frozen features from pre-training. LARS~\cite{you2017scaling} is used as the optimizer with weight decay as $0$ and momentum as $0.9$. The base learning rate is $0.02$ with $4096$ batch size, and a cosine decay schedule is used for 90 epochs.

\section{Additional Results}
\label{appendix:results}


\subsection{Projector Structure}
The design of projector is another main factor that influences the final performance and also varies across different works.
\cite{he2020momentum} applies linear projection to contrastive learning. SimCLR~\cite{chen2020simple} finds that a 2-layer MLP can help boost the performance. SimSiam~\cite{chen2021exploring} further extends the projector depth to 3. We explore the effects of different projector depths in Table~\ref{tab:pr_layers}. Here \name\ with 100 epochs pre-training is used. The results show that increasing the depth of projector from $1$ to $3$ can greatly boost the linear evaluation accuracy. However, the improvement saturates when the projector becomes deeper.

Moreover, Barlow Twins~\cite{zbontar2021barlow} extends the dimension of projector from 2048 to 8192, showing notable improvement. We further study the effect of projector's width in Table~\ref{tab:pr_dims}. For simplicity, we change the output dimension together with the hidden dimension. \name\ with 100 epochs pre-training is used. It's shown that increasing the projector width can steadily increase the performance, and does not seem to saturate even the dimension is increased to $16384$. 
\begin{table*}[t]
    \centering
    \begin{tabular}{@{}cccccccccc@{}}
    \toprule
    \multirow{2}{*}{Method} &  \multicolumn{3}{c}{VOC07+12 detection}  & \multicolumn{3}{c}{COCO detection} & \multicolumn{3}{c}{COCO instance seg} \\
    & AP$_\mathrm{all}$ & AP$_{50}$ & AP$_{75}$ & AP$_\mathrm{all}^{\mathrm{box}}$ & AP$_{50}^{\mathrm{box}}$ & AP$_{75}^{\mathrm{box}}$ & AP$_\mathrm{all}^{\mathrm{mask}}$ & AP$_{50}^{\mathrm{mask}}$ & AP$_{75}^{\mathrm{mask}}$\\
    \midrule
    Supervised & 54.7 & 84.5 & 60.8 & 38.9 & 59.6 & 42.7 & 35.4 & 56.5 & 38.1 \\
    MoCov2~\cite{chen2020improved} & 56.4 & 81.6 & 62.4 & 39.8 & 59.8 & 43.6 & 36.1 & 56.9 & 38.7  \\
    SimCLR~\cite{chen2020simple} & 58.2 & 83.8 & 65.1 & 41.6 & 61.8 & 45.6 & 37.6 & 59.0 & 40.5 \\ 
    DINO~\cite{caron2021emerging} & 57.2 & 83.5 & 63.7 & 41.4 & 62.2 & 45.3 & 37.5 & 58.8 & 40.2  \\
    TWIST~\cite{wang2021self} & 58.1 & 84.2 & 65.4 & 41.9 & 62.6 & 45.7 & 37.9 & 59.7 & 40.6 \\
    \name & 57.8 & 84.0 & 64.9 & 42.0 & 62.6 & 45.7 & 37.9 & 59.7 & 40.7 \\
    \bottomrule
    \end{tabular}
    \caption{Transfer learning: object detection and instance segmentation. VOC benchmark uses Faster R-CNN with FPN. COCO benchmark uses Mask R-CNN with FPN. The supervised VOC results are run by us.}
    \label{tab:det}
\end{table*}

\begin{table}
  \centering
  \begin{tabular}{@{}cccccc@{}}
    \toprule
     Depth     & 1 & 2 & 3 & 4 & 5 \\
     \midrule
     Linear Eval    & 60.7 & 65.2 & 70.3 & 70.0 & 69.8 \\ 
    \bottomrule
  \end{tabular}
  \caption{Effect of projector depth.}
  \label{tab:pr_layers}
\end{table}

\begin{table}
  \centering
  \begin{tabular}{@{}cccccc@{}}
    \toprule
    Width & 1024 & 2048 & 4096 & 8192 & 16384 \\
    \midrule
    Linear Eval & 68.3 & 70.3 & 70.5 & 70.9 & 71.2 \\ 
    \bottomrule
  \end{tabular}
  \caption{Effect of projector width.}
  \label{tab:pr_dims}
\end{table}

\begin{table}
    \centering
    \begin{tabular}{@{}lcccc@{}}
    \toprule
    \multirow{2}{*}{Method} &  \multicolumn{2}{c}{1\%}  & \multicolumn{2}{c}{10\%} \\
    & Top 1 & Top 5 & Top 1 & Top 5\\
    \midrule
    Supervised & 25.4 & 48.4 & 56.4 & 80.4  \\
    SimCLR~\cite{chen2020simple} & 48.3 & 75.5 & 65.6 & 87.8  \\ 
    BYOL~\cite{jean-bastien2020bootstrap} & 53.2 & 78.4 & 68.8 & 89.0 \\
    Barlow Twins~\cite{zbontar2021barlow} & 55.0 & 79.2 & 69.7 & 89.3 \\
    DINO~\cite{caron2021emerging} & 52.2 & 78.2 & 68.2 & 89.1 \\
    TWIST~\cite{wang2021self} & 61.2 & 84.2 & 71.7 & 91.0 \\
    \name & 60.8 & 83.8 & 71.5 & 90.6 \\
    \bottomrule
    \end{tabular}
    \caption{Semi-supervised learning on ImageNet.}
    \label{tab:semisup}
\end{table}

\subsection{Semi-supervised Learning}
We finetune the pretrained model on the 1\% and 10\% subset of ImageNet's training set, following the standard protocol in \cite{chen2020simple}. The results are reported in Table~\ref{tab:semisup}. Compared with previous methods, \name\ is able to obtain comparable results with \cite{wang2021self} and obtain 5\% and 1\% improvement from other methods on the 1\% and 10\% subset, respectively.

\subsection{Transfer Learning}
We also transfer the pretrained model to downstream stasks, including PASCAL VOC~\cite{everingham2010pascal} object detection, COCO~\cite{lin2014microsoft} object detection and instance segmentation. The model is finetuned in an end-to-end manner. Table~\ref{tab:det} shows the final results. It can be seen that \name\ delivers competitive transfer performance with other self-supervised learning methods, and surpasses the supervised baseline.

\section{Licenses of Assets}
\textbf{ImageNet}~\cite{deng2009imagenet} is subject to the ImageNet terms of access~\cite{imagenet_terms}.

\textbf{PASCAL VOC}~\cite{everingham2010pascal} uses images from Flickr, which is subject to the Flickr terms of use~\cite{fickr}.

\textbf{COCO}~\cite{lin2014microsoft}. The annotations are under the Creative Commons Attribution 4.0 License. The images are subject to the Flickr terms of use~\cite{fickr}.

\end{document}